\documentclass[preprint,12pt]{elsarticle}

\usepackage{amssymb}
\usepackage{amsmath}
\usepackage{multirow}
\usepackage{colortbl}
\usepackage{xcolor}
\usepackage{graphicx}
\definecolor{mygrey}{gray}{0.9}

\journal{Neurocomputing}

\begin{document}

%% Separate Abstract Page (First Page)
\noindent\textbf{\large Abstract:}
\vspace{1em}

\noindent\textbf{Data Augmentation for High-Fidelity Generation of CAR-T/NK Immunological Synapse Images}

\noindent Xiang Zhang, Boxuan Zhang, Alireza Naghizadeh, Mohab Mohamed, Dongfang Liu*, Ruixiang Tang*, Dimitris Metaxas*, Dongfang Liu*
\vspace{1em}

\noindent Chimeric antigen receptor (CAR)-T and NK cell immunotherapies have transformed cancer treatment, and recent studies suggest that the quality of the CAR-T/NK cell immunological synapse (IS) may serve as a functional biomarker for predicting therapeutic efficacy. Accurate detection and segmentation of CAR-T/NK IS structures using artificial neural networks (ANNs) can greatly increase the speed and reliability of IS quantification. However, a persistent challenge is the limited size of annotated microscopy datasets, which restricts the ability of ANNs to generalize. To address this challenge, we integrate two complementary data-augmentation frameworks. First, we employ Instance Aware Automatic Augmentation (IAAA), an automated, instance-preserving augmentation method that generates synthetic CAR-T/NK IS images and corresponding segmentation masks by applying optimized augmentation policies to original IS data. IAAA supports multiple imaging modalities (e.g., fluorescence and brightfield) and can be applied directly to CAR-T/NK IS images derived from patient samples. In parallel, we introduce a Semantic-Aware AI Augmentation (SAAA) pipeline that combines a diffusion-based mask generator with a Pix2Pix conditional image synthesizer. This second method enables the creation of diverse, anatomically realistic segmentation masks and produces high-fidelity CAR-T/NK IS images aligned with those masks, further expanding the training corpus beyond what IAAA alone can provide. Together, these augmentation strategies generate synthetic images whose visual and structural properties closely match real IS data, significantly improving CAR-T/NK IS detection and segmentation performance. By enhancing the robustness and accuracy of IS quantification, this work supports the development of more reliable imaging-based biomarkers for predicting patient response to CAR-T/NK immunotherapy.

\newpage

\begin{frontmatter}

%% Title
\title{Data Augmentation for High-Fidelity Generation of CAR-T/NK Immunological Synapse Images}

\author[rit]{Xiang Zhang}
\ead{xz2649@rit.edu}

\author[rutgers]{Boxuan Zhang}
\ead{boxuan.zhang@rutgers.edu}

\author[njms]{Alireza Naghizadeh}
\ead{ar.naghizadeh@rutgers.edu}

\author[njms]{Mohab Mohamed}
\ead{mm2233@scarletmail.rutgers.edu}

\author[rit]{Dongfang Liu\corref{cor}}
\ead{dxleec@rit.edu}

\author[rutgers]{Ruixiang Tang\corref{cor}}
\ead{ruixiang.tang@rutgers.edu}

\author[rutgers]{Dimitris Metaxas\corref{cor}}
\ead{dnm@cs.rutgers.edu}

\author[njms]{Dongfang Liu\corref{cor}}
\ead{dongfang.liu@rutgers.edu}

\cortext[cor]{Corresponding author}

\affiliation[rit]{organization={Rochester Institute of Technology},
            city={Rochester},
            state={NY},
            country={USA}}

\affiliation[rutgers]{organization={Rutgers University},
            city={New Brunswick},
            state={NJ},
            country={USA}}

\affiliation[njms]{organization={Rutgers New Jersey Medical School},
            city={Newark},
            state={NJ},
            country={USA}}

%% Abstract
\begin{abstract}
Chimeric antigen receptor (CAR)-T cell immunotherapies have transformed cancer treatment, and recent studies suggest that the quality of the CAR-T/NK cell immunological synapse (IS) may serve as a functional biomarker for predicting therapeutic efficacy. Accurate detection and segmentation of CAR-T/NK IS structures using artificial neural networks (ANNs) can greatly increase the speed and reliability of IS quantification. However, a persistent challenge is the limited size of annotated microscopy datasets, which restricts the ability of ANNs to generalize. To address this challenge, we integrate two complementary data-augmentation frameworks. First, we employ Instance Aware Automatic Augmentation (IAAA), an automated, instance-preserving augmentation method that generates synthetic CAR-T/NK IS images and corresponding segmentation masks by applying optimized augmentation policies to original IS data. IAAA supports multiple imaging modalities (e.g., fluorescence and brightfield) and can be applied directly to CAR-T/NK IS images derived from patient samples. In parallel, we introduce a Semantic-Aware AI Augmentation (SAAA) pipeline that combines a diffusion-based mask generator with a Pix2Pix conditional image synthesizer. This second method enables the creation of diverse, anatomically realistic segmentation masks and produces high-fidelity CAR-T/NK IS images aligned with those masks, further expanding the training corpus beyond what IAAA alone can provide. Together, these augmentation strategies generate synthetic images whose visual and structural properties closely match real IS data, significantly improving CAR-T/NK IS detection and segmentation performance. By enhancing the robustness and accuracy of IS quantification, this work supports the development of more reliable imaging-based biomarkers for predicting patient response to CAR-T/NK immunotherapy.
\end{abstract}


\begin{highlights}
\item We propose IAAA for instance-aware data augmentation with accurate masks.
\item We introduce SAAA using diffusion and Pix2Pix for scalable image synthesis.
\item Both methods significantly improve CAR-T/NK cell detection and segmentation.
\item IAAA preserves instance fidelity; SAAA enables unlimited data generation.
\end{highlights}

%% Keywords

\end{frontmatter}



\section{Introduction}

In translational and basic biomedical research, well-annotated datasets are critical for machine learning algorithms. Particularly, the detection and segmentation of nuclei cells used for intensity quantification of these cells require large manually annotated microscopical datasets for artificial neural networks (ANNs)~\cite{mehlig2021machine,wangGeneSegNetDeepLearning2023,10.1093/bib/bbae407}. The task of object detection aims to grasp a visually strong understanding by estimating the locations of varying concepts found within the image~\cite{jiao2019survey}. This fundamental computer vision problem is recently approached through ANNs, which classically require an abundance of training data~\cite{zhao2019object,tan2020efficientdet}.

However, the accessibility of varied datasets with thousands of correctly segmented images is infeasible for many fields of medical imaging. To solve these problems, the current trend is to achieve competitive results on smaller datasets by using innovative techniques such as data augmentation~\cite{liu2019generative,rahman2020any,brigato2021close}. Data augmentation allows for the extrapolation of small datasets to improve a deep-learning model~\cite{rizk2019effectiveness,clark2021training}. The performance of neural networks often improves with the amount of data available, allowing the network to learn from different variations of the training set. Classic data augmentation techniques include image operations such as cropping, padding, and horizontal flipping.

The other major methods for augmenting data are Generative adversarial networks (GANs). These methods approach the task of artificial image generation through a unique framework consisting of two models, a generator, and a discriminator~\cite{goodfellow2014generative}. The generator creates fake data using the feedback outputted by the discriminator, which is a classifier that distinguishes the generated data from the real data.

However, current augmentation methods do not create accurate masks for objects and lack the information that is required for detection tasks for instance segmentation and bounding box detection algorithms. Many of these methods also have other problems that make them unsuitable for generating training data in basic research field, such as the requirement for large sets of training data and restriction on output resolution. There are several attempts to solve the lack of training data~\cite{gurumurthy2017deligan,noguchi2019image,wang2018transferring,wang2020minegan,zhang2017stackgan}. DeLiGan parametrizes the latent generative space and learns the model's parameters along with those of GAN~\cite{gurumurthy2017deligan}. Noguchi and Wang apply the principles of transfer learning to a pre-trained generator to compensate for the lack of a larger dataset~\cite{noguchi2019image}. Another approach is to directly apply commonly used augmentation techniques such as cropping, flipping, scaling, color jittering, and region masking~\cite{inoue2018data,ho2019population,devries2017dataset,krizhevsky2017imagenet} on the training data to increase its variety.

A fundamental challenge with applying standard data augmentation techniques to nuclei and cell microscopy images arises from the inherently sensitive nature of biomedical data. Medical imaging supports high-stakes prediction tasks where noisy, distorted, or biologically implausible synthetic samples can degrade model performance and lead to unreliable outcomes. In contrast to natural image generation, where creating visually convincing but semantically ambiguous objects may be acceptable, biomedical applications require strict preservation of anatomical structure and cellular morphology. This requirement is especially important in CAR-T/NK research, where the quality of the CAR immunological synapse (IS) has been proposed as a functional biomarker for predicting therapeutic efficacy. However, the scarcity of annotated CAR IS datasets has limited the broader development of ANN-based IS analysis tools. To mitigate this limitation, we introduce Instance Aware Automatic Augmentation (IAAA) for cell nuclei microscopy images. Our objective is to generate artificial images paired with accurate instance segmentation masks that improve object detection and instance segmentation performance. IAAA applies augmentation techniques in a controlled manner that preserves the authenticity and structural integrity of cellular objects, with particular focus on IS morphology.

In addition to IAAA, we incorporate a complementary Semantic-Aware AI Augmentation (SAAA) generation pipeline that greatly expands dataset diversity while maintaining strict semantic and morphological fidelity. This pipeline combines an unconditional diffusion-based mask generator capable of producing anatomically plausible segmentation masks with an enhanced Pix2Pix conditional image synthesis model that translates these masks into high-resolution, biologically realistic microscopy images. By jointly modeling cell shape distributions and pixel-level appearance, the pipeline produces large numbers of realistic single-cell and multi-cell configurations that enrich structural variability, preserve fine cellular boundaries, and improve the robustness of downstream detection and segmentation models.

Together, these approaches enable the construction of instance-aware datasets that offer both high quality and detailed structural information needed for biomedical vision tasks. These methods are effective because cell nuclei microscopy images typically contain well-isolated cellular structures against relatively uniform background patterns, making controlled augmentation and synthetic generation feasible and biologically reliable.

The two augmentation frameworks serve complementary roles. IAAA operates directly on real images, applying optimized transformation policies to existing annotated samples while preserving instance-level accuracy; it is particularly effective when high-fidelity masks are already available. SAAA, by contrast, generates entirely new mask-image pairs from learned distributions, enabling dataset expansion beyond the original sample pool. By combining both approaches, we leverage the precision of instance-aware augmentation together with the diversity of generative synthesis.

The paper is organized as follows. \S\ref{sec:methods} introduces the technical background and details of the IAAA pipeline. \S\ref{sec:results} presents the experimental setup and evaluation metrics. \S\ref{sec:saaa} extends IAAA with diffusion-based mask generation and semantic synthesis (SAAA). Finally, \S\ref{sec:discussion} concludes with a discussion of limitations and future directions.

\section{Methods}\label{sec:methods}

\subsection{Greedy AutoAugment}
Greedy AutoAugment is an efficient search algorithm that finds the best augmentation policies within an arbitrarily large sample space. In this section, we discuss the search space, score, and search algorithm in more detail.

\paragraph{Search Space}
The search space, $S$, consists of $m$ sequential image operations. Each image operation can be defined as a sub-policy that contains information on two hyperparameters, the probability of applying the operation, and its magnitude. The range of probabilities and magnitude include two discrete variables within uniform spaces, $n_p$, and $n_m$. In this manner, a discrete search algorithm may be utilized to find the best sub-policies. The size of the search space for $n_o$ operations can be denoted as
\begin{equation}
n_s = (n_o\, n_p\, n_m)^f.
\end{equation}

\paragraph{Score}
A score is given to each sub-policy through the utilization of an autoencoder that compares how well an augmented cell maps to its original variant. Specifically, in our framework, the score is a metric that measures the performance of a policy by passing a given cell through a trained Wasserstein AutoEncoder. The degree of accuracy provided by this method measures the closeness of an image to the original cell dataset and is passed to the search algorithm to select the strongest augmentation policies.

\paragraph{Search Algorithm}
The size of the search space of potential augmentation policies grows exponentially due to the different combinations of varying transformations, their probability of applying a policy, and their magnitude. This explains the impracticality of a brute-force approach. To make the search process feasible, a greedy search is utilized. A reduced search space is traversed where each augmentation policy contains only a single sub-policy in the beginning, $l = 1$. In this reduced space, we find the best probability and magnitude for each of the varying image operations through the scoring criterion mentioned earlier. Similarly, we find the best variables for a second sub-policy, $l = 2$. For every found policy within the first stage, we find the best combinations of image operations with their associated probabilities and magnitudes. This process is repeated until $l = l_{\max}$, the maximum number of sub-policies. All policies are then sorted via their scores, and the top $b$ augmentations are selected to create the artificial training set.

\paragraph{Wasserstein AutoEncoder}
{To score each policy, our method employs a Wasserstein AutoEncoder (WAE)~\cite{tolstikhin2017wasserstein} trained to learn an identity function $f$ that maps an image $S_i$ from a set $S$ to itself. The WAE learns the distribution of real images and measures how closely an augmented cell resembles the original data by minimizing the Wasserstein distance between two probability distributions~\cite{tolstikhin2017wasserstein}. This scoring criterion ensures that augmentation policies preserve the biological authenticity of the cells~\cite{chong2020effectively}.

\subsection{Artificial Background Images}

To create natural background images, we take advantage of the images acquired from the existing training datasets. In these images, the existing cells need to be removed and then replaced with similar pixels that replicate the textures of the images. To remove the cells, first, manually segmented areas are used to create bounding boxes around the cells. If the bounding boxes are interconnected, a bigger bounding box is created to contain all the connected boxes. Next, the pixels inside each bounding box (inside pixels) need to be replaced by other pixels found outside of the bounding box (outside pixels) that are similar to the bordering pixels of the bounding box. For each inside pixel, $n$ outside pixels are selected randomly and compared to the pixels of the upper-left corner, upper-right corner, lower-left corner, and lower-right corner of the bounding box.

Among the selected pixels, the outside pixel that is most similar to the mentioned corners of the bounding box (evaluated using the mean Euclidean distance) is used to replace the inside pixel. This process is repeated until all inside pixels are replaced with outside pixels. Through this method, cells are removed from the bounding boxes with a similar texture to the original background. However, there would still be notable discrepancies between the background and transformed bounding boxes. To impede these problems and create smoother textures, we use a masking technique that utilizes a Gaussian filter to replace the previous bounding box with a new one.

In this method, an image of a similar size to the bounding box is created with a black-colored background and a white rectangular area in the middle that acts as the mask. Next, a Gaussian filter is applied to disperse the white rectangle based on a Gaussian distribution. The filtering process aids in a smooth transition between the textures of the background and the newly created bounding box. The new mask is finally used to copy the newly created bounding box onto the image. By repeating this process for each segmented cell, all the cells are removed from the image.

This process is shown in Fig.~\ref{fig:1}. Fig.~\ref{fig:1}a shows a sample from the Neural dataset in its original form with five cells that need to be removed. In Fig.~\ref{fig:1}a--e, red arrows show which cell will be removed next. Fig.~\ref{fig:1}b shows the process of the removal of the first cell. At the top of each image are small rectangles that show the process needed to remove the cells and replace them with a smooth texture. From left to right, we see a newly created bounding box, the preliminary mask, a filtered version of the mask using a Gaussian filter, and an applied filtered mask on the new bounding box. From the image, we see that the removal process is effective, and the removed area is not distinguishable from the rest of the image. From Fig.~\ref{fig:1}c to Fig.~\ref{fig:1}f we see how the other cells are removed. The newly generated background can now be used to place augmented cells to create artificial images.

\subsection{Placement of Cells on the Background}

The placement of cells on a background shares a similar problem to the removal of cells. The brighter color spectrum found in the background and the cells signifies the differences in textures between them, which results in unnatural-looking images. To solve this problem, first, manually segmented areas are used to create bounding boxes around each cell. If the bounding boxes are interconnected, a larger bounding box is generated to contain all the connected boxes. Next, we search for areas in the image that most closely resemble the pixel colors of the generated bounding boxes.

For each bounding box, $n$ number of $(x,y)$ points are randomly selected. To select the best $(x,y)$ pair, the upper-left corner, upper-right corner, lower-left corner, and lower-right corner of the bounding box are compared with $(x,y)$, $(x+w,y)$, $(x,y+h)$ pixels, where $w,h$ are the width and height of the bounding box. Among the selected $(x,y)$ pairs, the pair that is the most similar to the mentioned corners of the bounding box (evaluated using the mean Euclidean distance) is used to position the bounding box.

Finding a place where two textures have similar colors is helpful, but it is not enough, and there will still be notable discrepancies between the backgrounds and the newly placed cells. To solve this problem and create a smooth transition between the cell and the background, we use the same Gaussian filter technique. To place a cell in a background image, first, an image with a similar size to the bounding box is created with a black background and a white area in the middle that acts as a mask. Manual segmentation of the cell is then used to determine the shape of the white area.
Next, a Gaussian filter is applied to disperse the white segmented area through a Gaussian distribution. The filtering process aids in a smooth transition between the texture of the background and the segmented cell. The new mask is then used to copy the separated bounding box from the cell onto the image. By repeating this process for all the segmented cells, any desired number of cells can potentially be placed on the image.

This process is shown in Fig.~\ref{fig:1}. In Fig.~\ref{fig:1}g--k, red arrows show where the next cell will be added. Fig.~\ref{fig:1}g shows the artificially generated background created in Fig.~\ref{fig:1}a--f. Fig.~\ref{fig:1}h shows the process of adding the first cell to the background. From the top of the image, each small rectangle shows the process needed to add a cell with a smooth transition. From left to right, we see the original bounding box of the cell, the new bounding box taken from the new location of the cell, a filtered version of the manually segmented mask using a Gaussian filter, and an applied filtered mask on a searched bounding box from the background. From Fig.~\ref{fig:1}h to Fig.~\ref{fig:1}l we see how other cells are added. The newly generated images can be used to train artificial neural networks. Note that for the segmentation images, the original unfiltered mask is used instead of Gaussian-filtered masks due to the Gaussian filter being less deterministic and may include areas other than the cells, which is not desired.

\begin{figure}[!t]
  \centering
  \includegraphics[width=\linewidth]{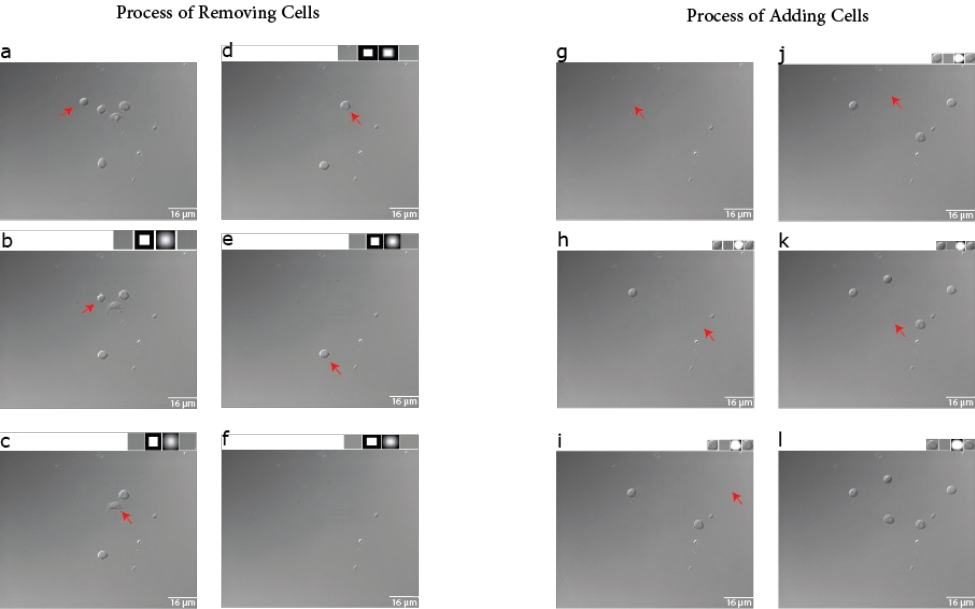}
  \caption{The removal of cells from a real image to create an artificially generated background and the placement of cells on the artificially generated background. (a) shows a sample of the real dataset from the Neural dataset. (b) shows the process of removing the first cell from the image. (c) (d) (e) (f) shows the process of removing four other cells. (g) shows an empty background. (h) shows the process of adding the first cell to the background. (i) (j) (k) and (l) show the process of adding four other cells.}
  \label{fig:1}
\end{figure}

\subsection{Structure of Instance Aware Automatic Augmentation (IAAA)}

IAAA categorizes segmented cells into Single-Cell Objects (SCOs), which are isolated well-defined cells, and Multi-Cell Objects (MCOs), which contain clustered cells with ambiguous boundaries. Fig.~\ref{fig:2}(a) shows the IAAA pipeline. The SCO \& MCO Generator extracts these cell objects from manually segmented images. Greedy AutoAugment~\cite{naghizadeh2020greedy,peng2018jointly,naghizadeh2021nshot,abavisani2020deep} identifies optimal augmentation policies scored by an autoencoder. The Image Generator applies these policies to cells and places them on empty backgrounds from the Background Generator (Fig.~\ref{fig:2}(b-c)).

As shown in Fig.~\ref{fig:2}(d), SCOs and MCOs are processed separately to preserve natural appearance. MCOs have ambiguous boundaries while SCOs are well-defined. When an MCO is placed inside an artificial image frame, it solely uses the MCO collection. Similarly, SCO placement uses only the SCO collection. Objects near image boundaries are excluded. The Image Generator mimics the training data distribution of cell counts and SCO/MCO ratios. For example, let $j$ denote the mode of the distribution representing the number of cells, then $j$ policies are selected and applied to $j$ original cells. These augmented cells are placed onto an empty background, avoiding overlaps while maintaining the ratio between SCOs and MCOs. This process generates $r$ artificial images with matching distributions and complete segmentation masks.

\begin{figure}[htbp]

  \centering
  \includegraphics[width=\linewidth]{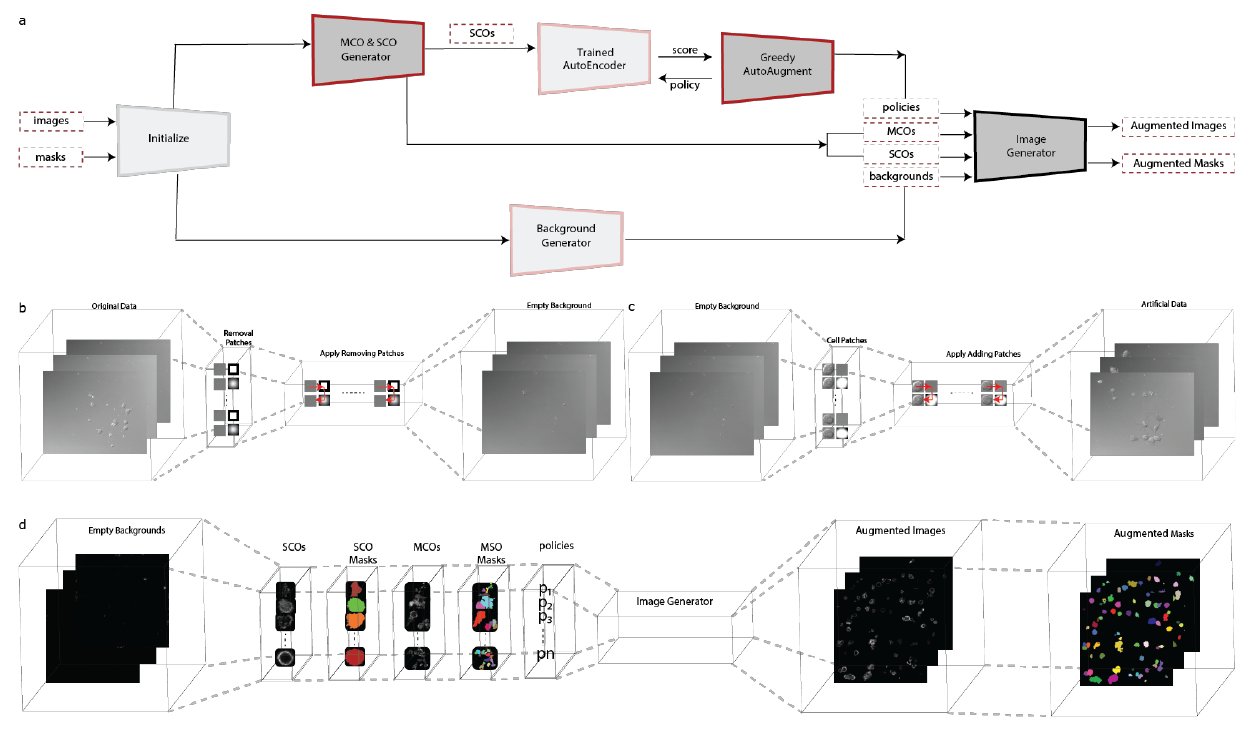}
  \vspace{-1em}
  \caption{Experimental diagram. (a) shows the overall structure of the IAAA technique. (b) Pipeline for creating background images. (c) Pipeline for adding cells on backgrounds. Gaussian filtering removes and adds patches seamlessly. (d) Pipeline for image generation.}
  \label{fig:2}
  \vspace{-2em}
\end{figure}

\section{Results}\label{sec:results}

\subsection{Evaluation Metrics}
To measure the quality of the images generated, we use two well-known evaluation metrics~\cite{kim2020paa}. Fr\'echet Inception Distance (FID) calculates the Fr\'echet distance between two multidimensional Gaussian distributions~\cite{kim2020paa}. This technique compares the distribution of the generated images, along with the mean and variances of the Gaussian distributions between artificial and original images. Kernel Inception Distance (KID) compares the two probability distributions by drawing samples independently from each distribution~\cite{kim2020paa}. This method improves FID and acts as a more reliable and unbiased estimator.

To measure the improvement in bounding box detection, we use variants of average precision (AP) to evaluate and compare how different object detectors perform given artificial data. Average precision computes the average precision value for recall values over $0$ to $1$. AP$_{50}$ and AP$_{75}$ are metrics used to evaluate the intersection over a union of objects (IoU). A perfect prediction yields an IoU value of $1$, while a completely wrong detection yields a value of $0$. A degree of overlap gives a value between the two. A decision can be made on how much overlap can potentially be considered a correct prediction. Hence, AP$_{50}$ and AP$_{75}$ give the average precision with IoU thresholds of $0.5$ and $0.75$, respectively. The AP$_s$ represent how well the model performs with small objects. Since AP$_s$ are used to evaluate the identification of small objects, it is specifically important for CAR-T/NK IS cell detection. All of the GAN models were trained for 150{,}000 epochs across all the experiments. For the deep neural networks in our method, we used PyTorch~\cite{paszke2019pytorch}. For WAE~\cite{tolstikhin2017wasserstein}, we used the implementation in~\cite{subramanian2020pytorchvae}.

\subsection{Data Preparation for Original Datasets}

We use two cell eukaryotic datasets containing microscopical images obtained from two different biological fields. The CAR-T/NK cell dataset consists of 156 images at a resolution of $1024 \times 1024$, divided into train, test, and validation sets with 93, 31, and 32 images. To test bright field images, we use the Neural dataset~\cite{kim2020paa} consisting of 644 images at a resolution of $640 \times 512$, also divided into train, test, and validation sets with 386, 129, and 129 images. Since the Neural dataset is based on DIC images, it shows the flexibility of the algorithm to handle different types of microscopical images, which is needed for automation of CAR-T/NK IS cells.

\subsection{Data Preparation for Mixed Datasets}

We mixed the images generated by our proposed method with the original images. The mixture is only for the training sets and not validation and test sets. After combining, the training set for the CAR-T/NK cell dataset contains 200 images, a combination of 93 original images and 107 generated images. The training set of the Neural dataset contains 800 images, a combination of 386 original images and 414 generated images.

\subsection{Evaluation of the Single-Cell Object (SCO) Image Quality Generation by Fr\'echet Inception Distance (FID) and Kernel Inception Distance (KID)}

SCOs are the most basic elements in the creation of complete artificial images. To create an artificial SCO, we use extracted segmentation masks manually. The cells should then transform from their original state to a new state using an augmentation policy. The outcome is an artificial cell that can be used to create artificial images. To assess the quality of the generated images, we follow a similar approach found in previous studies~\cite{chong2020effectively}.

The results of the image generation are presented in Fig.~\ref{fig:3}. Due to the simplified nature of SCO, most methods perform well in producing strong artificial SCO images. Fig.~\ref{fig:3}a--b shows a sample of SCO images and their associated masks that were generated by our method for the CAR-T/NK dataset. Respectively, Fig.~\ref{fig:3}c--d shows a sample of original SCO images with their associated masks. In all the images, the original images corresponded to the zoomed-in regions. From these images, the proposed method produces outputs with a great likeness to the original SCOs.

\begin{figure}[!t]
  \centering
  \includegraphics[width=\linewidth]{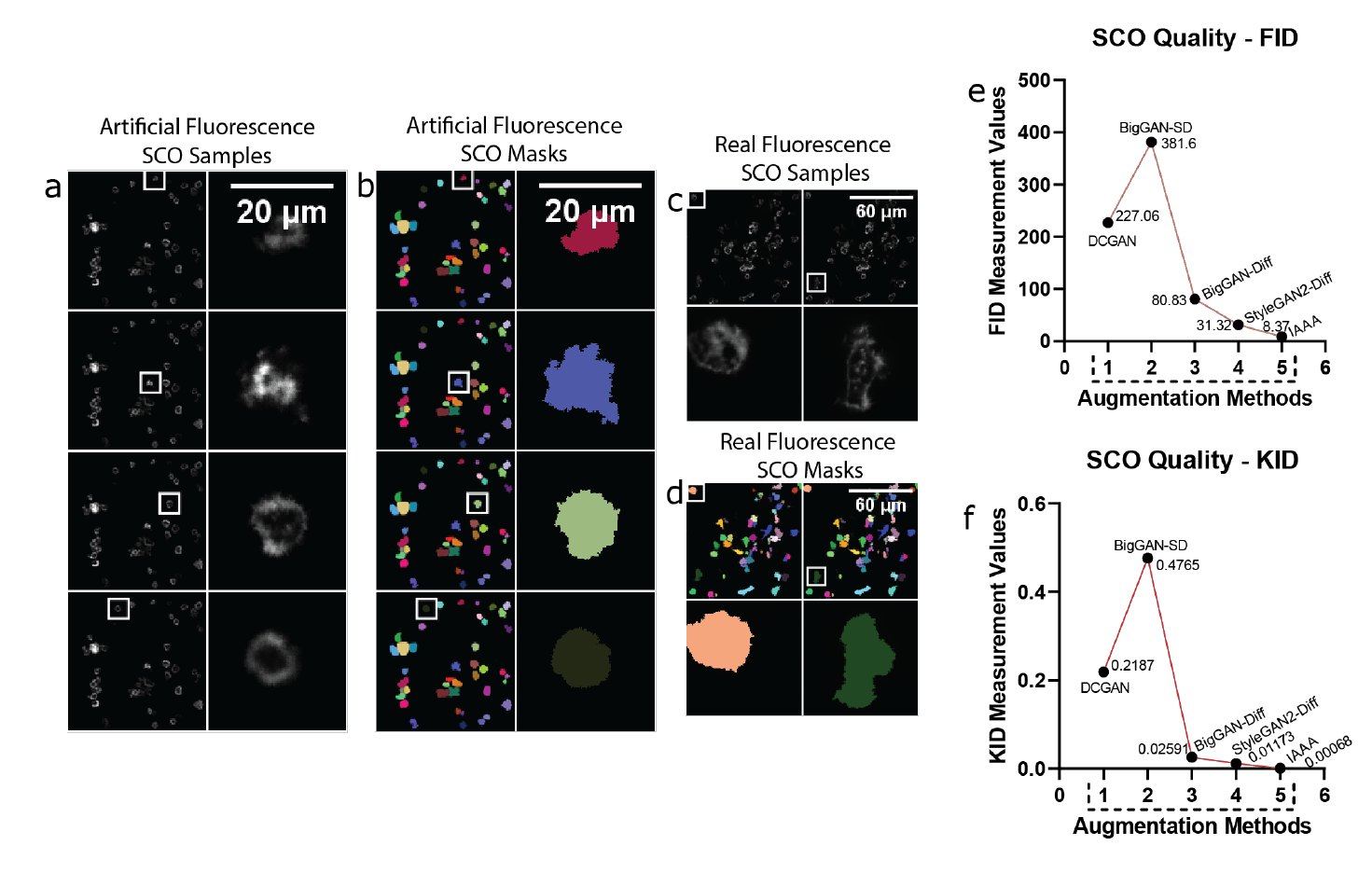}
  \caption{Representative image generation using SCOs. The (a) and (b) represent the artificial images and their masks generated by our method. The (c), and (d) represent reference (real) images which are accompanied by their associated masks. We zoom into different regions of the images for better visibility of SCOs. The FID and KID scores of SCOs, are presented in (e), and (f) using the CAR-T/NK dataset. For comparison, the quality of images from IAAA is measured against four GAN methods (DCGAN, BigGAN-SD, BigGAN-Diff, and StyleGAN2-Diff).}
  \label{fig:3}
\end{figure}

To evaluate the quality of the images, we use the Fr\'echet Inception Distance (FID) and the Kernel Inception Distance (KID), which have shown better support for smaller datasets~\cite{binkowski2018demystifying}. In other words, we answer how much applying different augmentation policies affects the quality of SCOs. The average scores are reported with numerical values and are computed over three random sets of sample generation.

For the FID scores (Fig.~\ref{fig:3}e), the results show that our method performs competitively compared to existing models. The best-performing GAN model is StyleGAN2-Diff, which is outperformed by 22.95 using the proposed method with the CAR-T/NK data. The KID scores (Fig.~\ref{fig:3}f) show that the proposed method has a higher score than StyleGAN2-Diff by 0.01105 using the CAR-T/NK data. This shows that the controlled environment was successful in the generation of SCOs and IAAA preserves the original traits of the natural cells found in the raw data.

\subsection{Evaluation of the Multi-Cell Object (MCO) Image Quality Generation by Fr\'echet Inception Distance (FID) and Kernel Inception Distance (KID)}

MCO is another basic element for eukaryotic cell microscopy images. The artificial generation of MCOs is more challenging for GAN models because their numbers are generally much lower than SCOs. This should not affect our model, which has minimal dependence on the quantity of images for the production of high-quality cells. Similar to the previous section, we investigate how the proposed method compares to GAN models used in the creation of realistic images.

The results to produce MCOs using the CAR-T/NK dataset are presented in Fig.~\ref{fig:4}. Similar to the SCO, samples are provided for observation. Fig.~\ref{fig:4}a--b shows a sample of MCO images and their associated masks that were generated by our method for the CAR-T/NK dataset. Respectively, Fig.~\ref{fig:4}c--d shows a sample of the original MCO images with their associated masks. In all the images, the original images corresponded to the zoomed-in regions. The results show that the controlled environment was again successful in generating new MCOs while simultaneously preserving the traits found within the original cells.

\begin{figure}[!t]
  \centering
  \includegraphics[width=\linewidth]{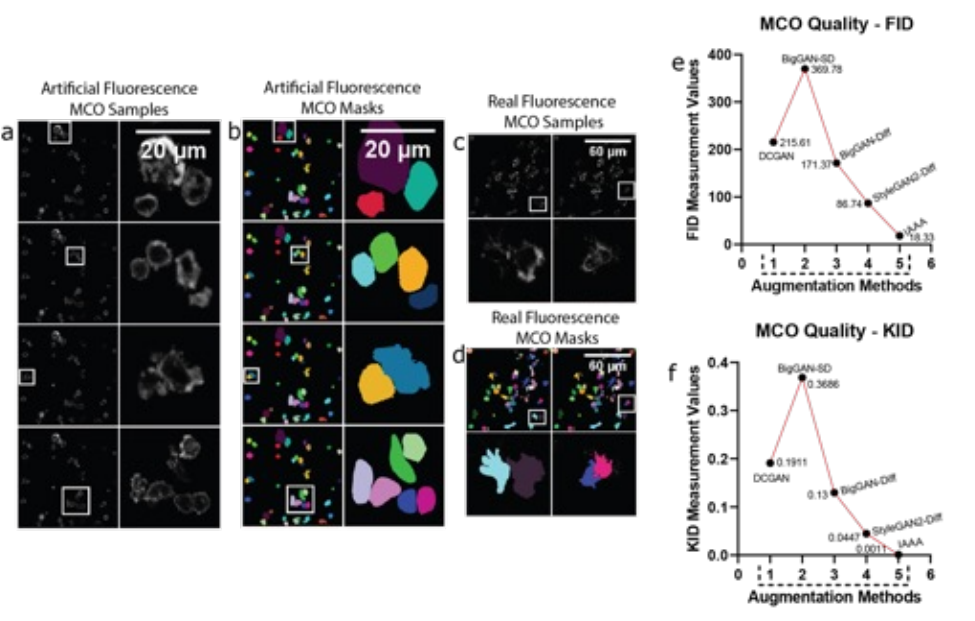}
  \caption{Representative image generation using MCOs. The (a) and (b) represent the artificial images and their masks generated by our method. The (c), and (d) represent reference (real) images which are accompanied by their associated masks. We zoom into different regions of the images for better visibility of MCOs. The FID and KID scores of MCOs are presented in (e), and (f) using the CAR-T/NK dataset. For comparison, the quality of images from IAAA is measured against four GAN methods (DCGAN, BigGAN-SD, BigGAN-Diff, and StyleGAN2-Diff).}
  \label{fig:4}
\end{figure}

For the FID scores (Fig.~\ref{fig:4}e), the results show that our method greatly outperforms the existing models across all parameters. The FID score of the proposed method surpasses the best-performing GAN model StyleGAN2-Diff by 68.41 within the CAR-T/NK dataset. For KID scores (Fig.~\ref{fig:4}f), the data shows the proposed method has a better generation quality for MCOs. The best performing GAN model is again StyleGAN2-Diff, which is outperformed by 0.04365 by our method within the CAR-T/NK dataset. The evaluation scores show the difficulty of generating naturalistic MCOs for GAN models due to the many variables that result in cell clustering and adhesion.

\subsection{Evaluation of the Quality of Final Image Generation by Fr\'echet Inception Distance (FID) and Kernel Inception Distance (KID)}

The proposed method utilizes two sets of images (SCOs and MCOs), along with recognized noisy backgrounds, to place them into single frames. When these elements are used to generate images, we create artificial images that, for the most part, are not distinguishable from natural data. Specifically, normal augmentation techniques are applied to the segmented SCOs and MCOs. To maintain the integrity of the original cells, only a subset of controlled augmentation techniques is used. To apply an augmentation onto the cells, we use an augmentation policy that consists of a set of different techniques at varying magnitudes. Search algorithms, such as Greedy AutoAugment~\cite{naghizadeh2020greedy,peng2018jointly,naghizadeh2021nshot,abavisani2020deep}, find the strongest augmentation policies in the search space of policies using a specified scoring criterion. The quantities of SCOs and MCOs are important for creating diverse images. Since we are utilizing small sets of training data, generating artificial data seems more difficult for GANs.

The results of the creation of complete CAR-T/NK cell images are presented in Fig.~\ref{fig:5}. We also provided more samples with higher resolution in the supplementary (Supplementary Figures S1--S19 show generated and real images, SCO and MCO samples, augmentation examples, and final image comparisons across the evaluated datasets). Fig.~\ref{fig:5}a--b shows a representative set of final images and their associated masks that were generated by our method for the CAR-T/NK dataset.

\begin{figure}[!t]
  \centering
  \includegraphics[width=\linewidth]{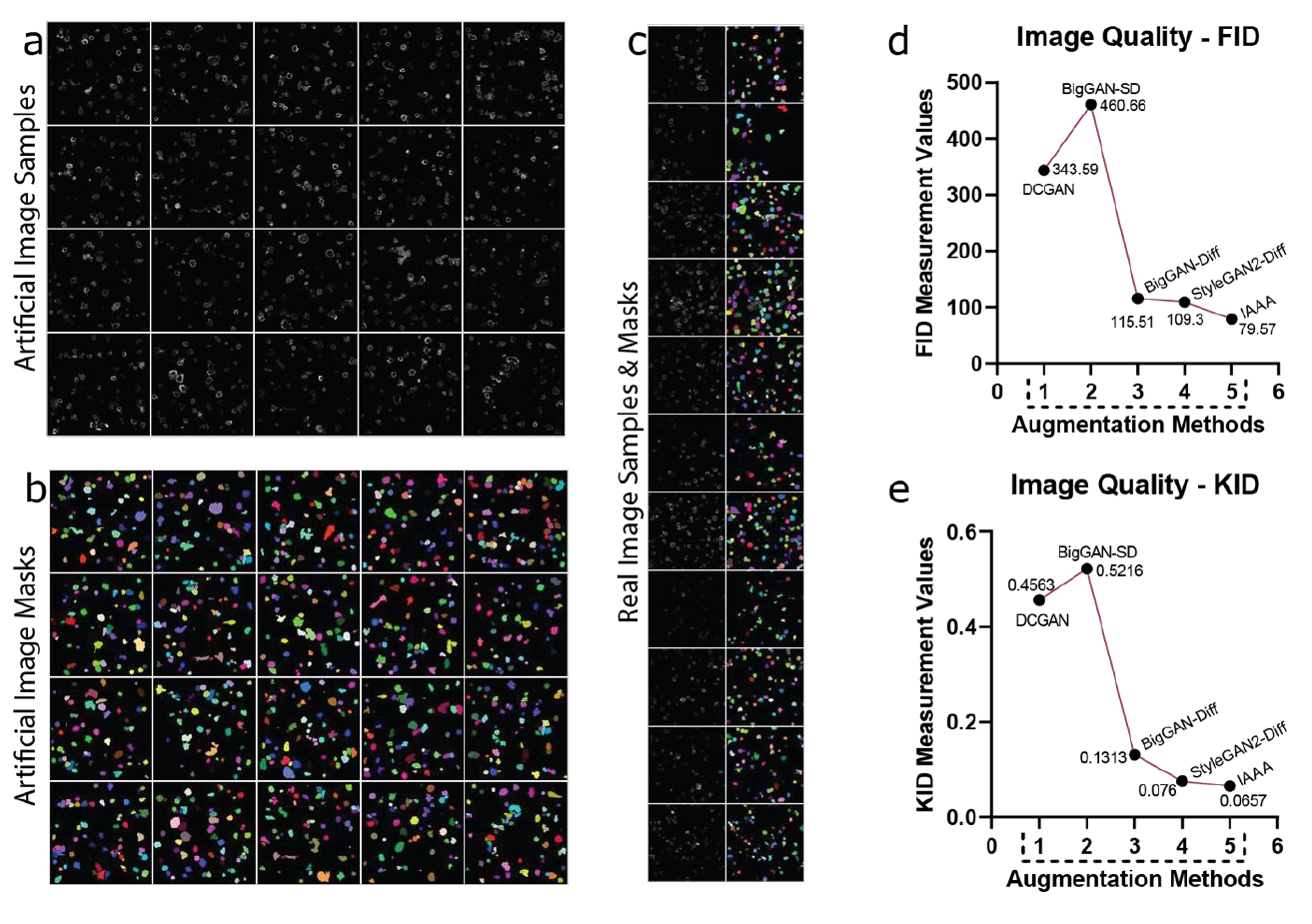}
  \caption{Representative image generation by MCOs using the CAR-T/NK dataset. The (a) and (b) represent the artificial images and their masks generated by our method. The (c) represents reference (real) images which are accompanied by their associated masks. The FID and KID scores are presented in (d), and (e) using the CAR-T/NK dataset. For comparison, the quality of images from IAAA is measured against four GAN methods (DCGAN, BigGAN-SD, BigGAN-Diff, and StyleGAN2-Diff).}
  \label{fig:5}
\end{figure}

Respectively, Fig.~\ref{fig:5}c shows a set of original images with their associated masks. In all the images, the original images are corresponded to the zoomed-in regions. The results show the precise segmentation masks that our method provides and demonstrate the capability to create high-quality images over many samples.

For the FID scores (Fig.~\ref{fig:5}d), the proposed method consistently outperformed the best-performing GAN method, Style- GAN2, by 29.73 with the CAR-T/NK dataset. The results remained consistent with the KID values (Fig.~\ref{fig:5}e), with the proposed method outperforming StyleGAN2 by 0.0103 using the CAR-T/NK dataset. The FID and KID scores demonstrate that similar to SCOs and MCOs, the IAAA could outperform GAN models to generate high-quality augmented images.

\subsection{Performance of CAR-T/NK IS Bounding Box Detection and Instance Segmentation}

This section shows the performance improvements of bounding box detection when the proposed augmentation method is applied. To measure the accuracy of cell detection, we use variants of average precision. For training, we follow the standard practice for bounding box detection algorithms and train for 12 epochs~\cite{zhao2020imageaugmentationsgan}. The experiments involve both multi-stage and one-stage bounding box detection algorithms.

We use the following state-of-the-art studies for bounding box detection, Faster R-CNN w/ FPN~\cite{lin2017fpn}, Cascade R-CNN~\cite{cai2019cascade}, Grid R-CNN~\cite{lu2019grid}, Libra R-CNN~\cite{pang2019libra}, RepPoints~\cite{yang2019reppoints}, FreeAnchor~\cite{zhang2019freeanchor}, FSAF~\cite{zhu2019fsaf}, ATSS~\cite{zhang2020atss}, PAA~\cite{kim2020paa}, and GFL~\cite{li2020generalized}. These networks are implemented on MMDetection software~\cite{chen2019mmdetection}, based on ResNet-101 (R-101) and ResNeXt-101 (X-101) backbones.
We investigate the method's effectiveness on the CAR-T/NK dataset (the first-row numbers in the BBox evaluation section in Table~\ref{tab:comparison}). According to the results, the augmented data consistently improves the results. For the AP scores, 11 out of 11 methods are improved. The AP$_{50}$ shows that 10 out of 11 methods show significant improvements, the AP$_{75}$ scores show that 11 out of 11 methods improve, and the AP$_s$ indicate that 9 out of 11 methods are improved. The improvement of the methods is consistent, with an average 6.32\% increase in overall AP evaluation across all the experiments.

For instance segmentation, we use data generated by our proposed method to improve cell segmentation accuracy and the robustness of model training results. To verify the effect of our method on CAR-T/NK IS images, the training results of the two datasets (Original Datasets and Mixed Datasets) are evaluated on the CAR-T/NK IS test sets for both datasets.

The results show that the method could easily improve the accuracy of object detection and instance segmentation of cells. The reason is that augmentation generates more data that benefits the generalization of the training process, which reduces overfitting and improves the robustness of the model. Our experiments show that there should be a balance between real and artificial samples. We observed that if the number of artificial samples is around the same number of real training data, it can generally increase the accuracy. For higher orders of artificial data, if the real data can provide high accuracy, overshadowing the data with artificial samples can have negative results. However, if the original data cannot provide high accuracy, higher artificial samples generally lead to better results.

\subsection{Generation of Artificial Bright-Field Microscopical Images using IAAA}
To test our method for bright-field microscopical images, we use the Neural dataset. The final results of creating complete Neural cell brightfield images are presented in Fig.~\ref{fig:6}. Fig.~\ref{fig:6}a-b shows the SCO images and their associated masks that were generated by our method for the Neural dataset. Fig.~\ref{fig:6}c-d shows the MCO images and their associated masks that were generated by our method for the Neural dataset. Similar to the CAR-T/NK results, Fig.~\ref{fig:6}e-f represents zoomed-in regions for SCOs and MCOs and samples from the original dataset.

In Fig.~\ref{fig:6}g--h, the average scores for FID and KID scores for this dataset are reported. The FID scores for DCGAN, BigGAN-SD, BigGAN-Diff, StyleGAN2-Diff, and IAAA are in order, 224.33, 278.45, 76.20, 37.73, and 52.49, respectively. The proposed method was able to create natural images and produce the second-best FID score when creating artificial images in $128\times128$ resolution. The results remained consistent for the KID values. The KID scores for DCGAN, BigGAN-SD, BigGAN-Diff, StyleGAN2-Diff, and IAAA are in order, 0.1803, 0.2361, 0.0619, 0.0147, and 0.0310. The proposed method also created natural images and produced the second-best KID score when creating artificial images in $128\times128$ resolutions.

\begin{figure}[!t]
  \centering
  \includegraphics[width=\linewidth]{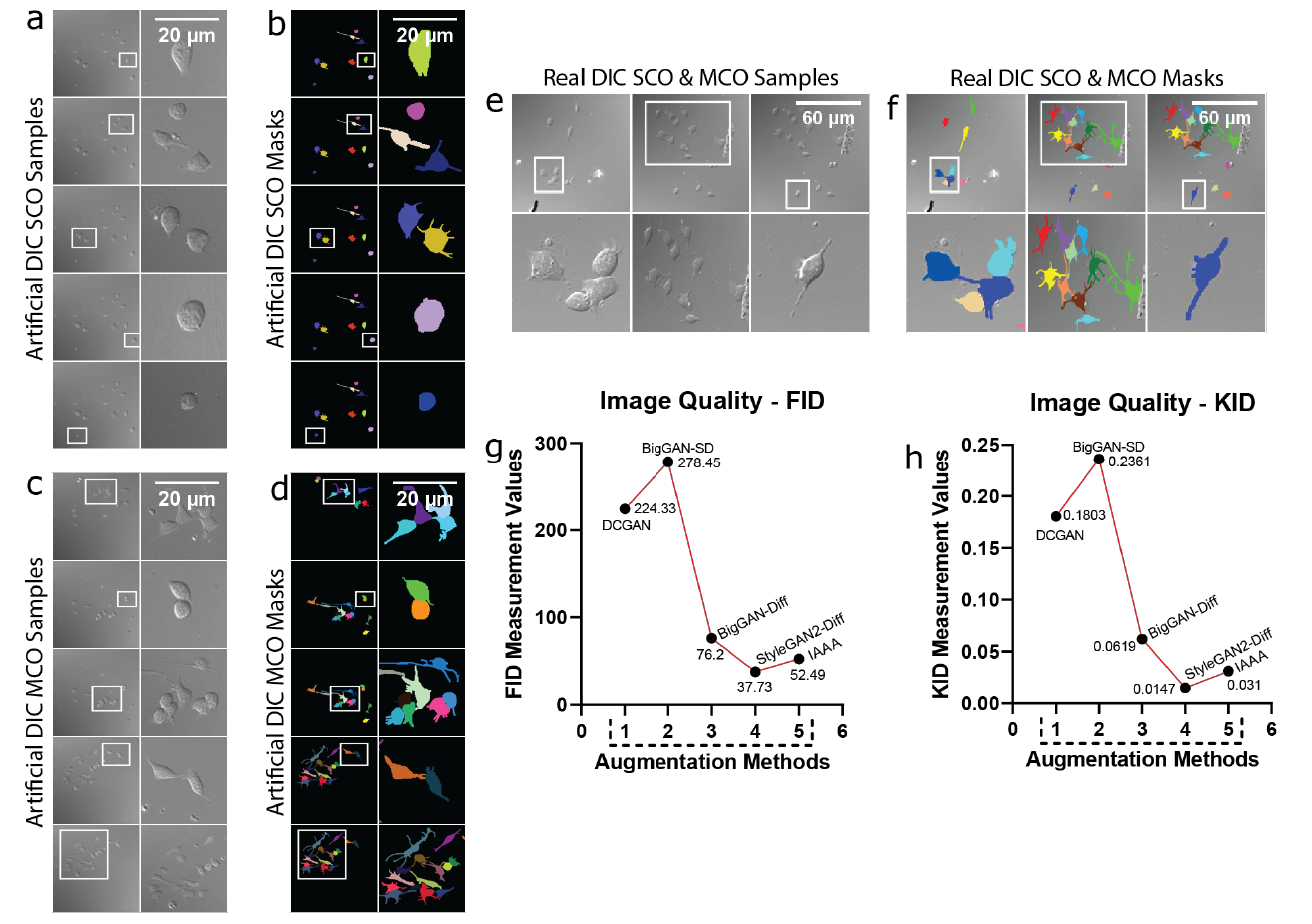}
  \caption{The final results of creating complete neural cell brightfield images. The (a), (b) (c), and (d) represent the artificial images and their masks generated by our method. The (e), and (f) represent reference (real) images which are accompanied by their associated masks. We zoom into different regions of the images for better visibility of SCOs and MCOs. The FID and KID scores of final images are presented in (g), and (h) using the Neural dataset. For comparison, the quality of images from IAAA is measured against four GAN methods (DCGAN, BigGAN-SD, BigGAN-Diff, and StyleGAN2-Diff).}
  \label{fig:6}
\end{figure}

\section{Extending IAAA with Diffusion-Based Mask Generation and Semantic Synthesis}\label{sec:saaa}

In addition to the instance-aware geometric augmentation provided by IAAA, many biomedical imaging settings still face a more fundamental limitation: the scarcity of high-quality segmentation masks themselves. IAAA relies on manually extracted SCOs and MCOs, which restricts its scalability when annotated masks are limited or labor-intensive to obtain. To overcome this bottleneck and further expand the diversity of training data, we introduce a Semantic-Aware AI Augmentation (SAAA) pipeline that synthesizes both new masks and corresponding microscopy images. In the SAAA method, an unconditional diffusion model learns the underlying distribution of cell shapes and produces diverse, biologically plausible segmentation masks, while a Pix2Pix-based semantic image generator translates these masks into realistic cell images. Together, this generative pipeline extends IAAA by enabling large-scale creation of structurally consistent image--mask pairs without requiring additional manual annotation, thereby enriching the dataset with controlled morphological variation and improved visual fidelity. The overall SAAA pipeline is illustrated in Fig.~\ref{fig:7}.

\begin{figure}[!t]
  \centering
  \includegraphics[width=\linewidth]{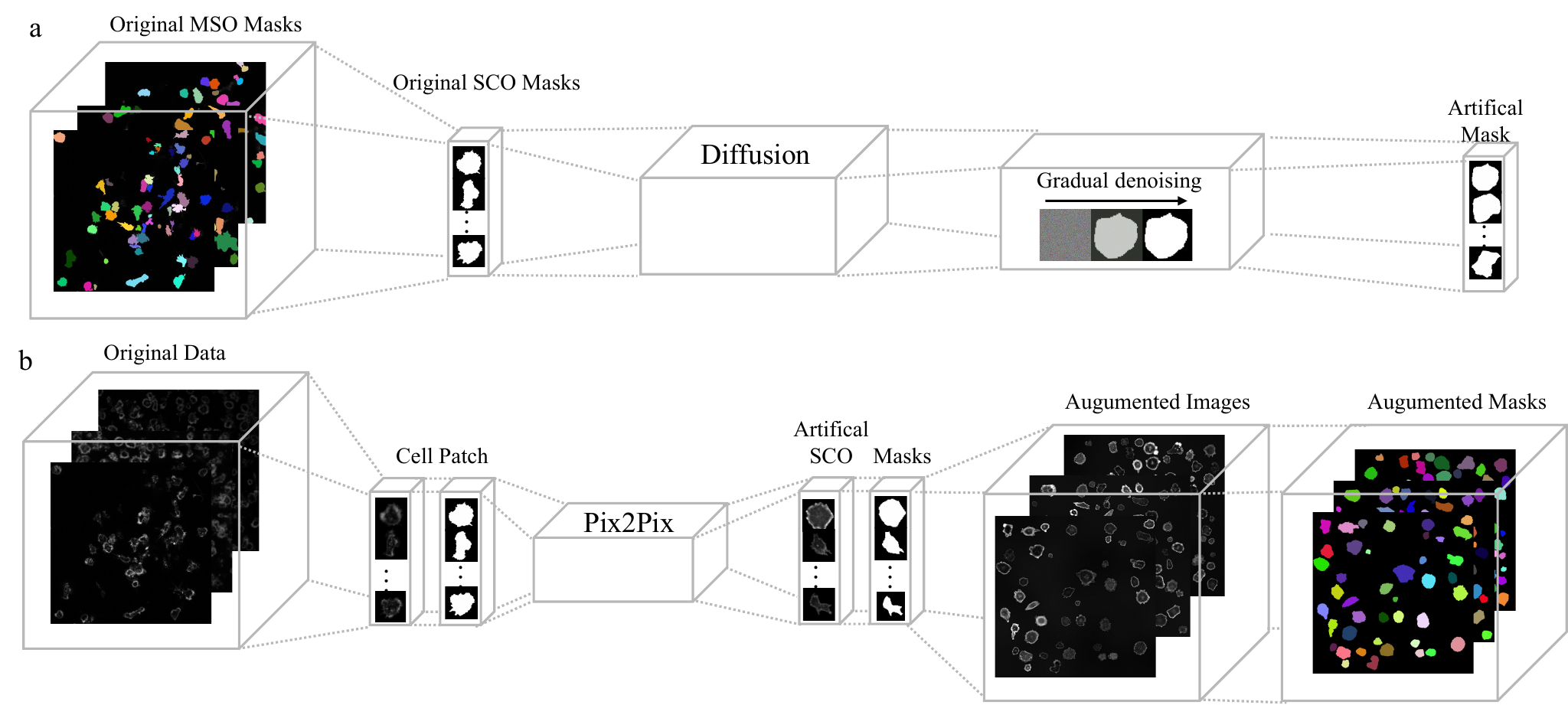}
  \caption{Experimental diagram of Semantic-Aware AI Augmentation. (a) Diffusion-based automated mask generation pipeline. The model is trained on SCO masks to learn the underlying distribution of cellular morphology, and after training it can generate an effectively unlimited number of new masks by denoising pure noise.  (b) Pix2Pix-based image synthesis pipeline. Pix2Pix learns the mapping between real cell masks and their corresponding microscopy images during training. At inference time, it takes either a true cell mask or a diffusion-generated mask as input and synthesizes a realistic cell image that matches the structure defined by the mask.}
  \label{fig:7}
\end{figure}

\subsection{Automated Mask Generator}

To address the limited availability of annotated segmentation masks, we introduce an automated mask generator based on an unconditional Denoising Diffusion Probabilistic Model (DDPM). Unlike conventional geometric augmentations such as flipping or rotation, which provide only modest variability, the diffusion model learns the underlying distribution of real cell masks and can therefore generate a broad range of novel, biologically realistic shapes. As shown in Fig.~\ref{fig:7}a, random noise is progressively denoised into cell-like structures that preserve essential morphological characteristics, including boundary smoothness and overall topology. This enables the creation of large numbers of high-quality synthetic masks from a small initial dataset, overcoming the quantitative bottlenecks inherent in manual annotation. These generated masks serve as inputs to the subsequent Pix2Pix-based semantic synthesis module.

\subsection{Semantic-Aware Image Generator Using Pix2Pix}

To translate the generated masks into realistic microscopy images, we employ a Pix2Pix-based conditional generative adversarial network (cGAN) that learns a direct mapping from segmentation masks to their corresponding cell images, as shown in Fig.~\ref{fig:7}b. Traditional cGAN architectures often face difficulties when applied to biomedical microscopy data, because the synthesized images may exhibit blurred or oversimplified textures, and the generated cell boundaries may fail to align precisely with the input masks. These limitations can reduce the biological integrity of the synthesized data and restrict its usefulness for downstream analysis.

To address these issues and to substantially enhance biological fidelity, we introduce several architectural improvements. First, we incorporate a Self-Attention module at the bottleneck of the network. This addition allows the generator to model long-range spatial relationships within the image, which is particularly important for multi-cell configurations where structural coherence across distant regions is necessary. Second, we integrate a Squeeze-and-Excitation (SE) block within the decoder. This component adaptively reweights channel-wise features, enabling the model to emphasize subtle but biologically meaningful patterns such as fine membrane textures or variations in fluorescence intensity. Third, we include a perceptual loss based on feature activations extracted from a pretrained VGG network. This perceptual term encourages the generator to produce images that exhibit realistic textures and consistent visual appearance, while also helping to mitigate boundary artifacts that commonly appear in naively trained cGANs.

The model is trained for 200 epochs using pairs of real microscopy images and their corresponding masks. The training set is further enriched with diffusion-generated masks to introduce a broader range of cell shapes and spatial arrangements. The optimization objective combines adversarial loss, L1 reconstruction loss, and the perceptual loss described above. Together, these terms guide the network to achieve both accurate structural alignment with the input masks and high fidelity in visual detail. After training, the Pix2Pix generator is capable of synthesizing high-quality single-cell and multi-cell microscopy images from diverse mask inputs, enabling large-scale dataset expansion and improving the robustness of downstream detection and segmentation models.

\subsection{Generation of Artificial Microscopical Images using SAAA}

To quantitatively evaluate the generated results, we use FID and KID following the same metrics described earlier. Together, these metrics capture both statistical realism and perceptual fidelity. We assess the performance of the Pix2Pix image generator. The CAR-T/NK dataset reached an FID of 28.7 and a KID of 0.025.

\subsection{Multi-cell Image Composition}

We analyzed the original dataset to estimate the average number of cells per image to simulate realistic spatial distributions. Then we randomly selected single-cell instances and composed multi-cell synthetic images based on the observed distribution. Each composite image was generated by placing a variable number of cells onto an empty background, guided by spacing constraints. The instance-level distinction was maintained by assigning unique colors to each mask. This process was repeated to produce a diverse set of large-scale synthetic images, which were subsequently used for downstream segmentation model training and evaluation.

\subsection{Bounding Box Detection Performance}

Following standard practice in MMDetection, we trained all detection models for 12 epochs and compared three settings: the original dataset and the IAAA and SAAA methods with 100 extra augmented samples applied to the original dataset. As shown in Table~\ref{tab:comparison}, the SAAA method consistently improves detection accuracy across both multi-stage and one-stage detectors. Averaged across all models, the SAAA training set yields performance gains of +2.01 AP, +0.73 AP$_{50}$, +1.80 AP$_{75}$, and +2.60 AP$_s$. It highlights the effectiveness of the SAAA method and its substantial improvements in detection accuracy.

\begin{table*}[htbp]
\centering
\caption{Comparison of detection performance between original methods, IAAA and SAAA. The Original column reports results obtained using only the original images, while the IAAA and SAAA columns show performance after adding 100 augmented images generated by the corresponding methods to the original training set. Detection accuracy is evaluated on 11 widely used multi-stage and one-stage detectors in MMDetection.}
\label{tab:comparison}
\renewcommand{\arraystretch}{1.1}
\resizebox{\textwidth}{!}{
\begin{tabular}{|l|l|c|cccc|cccc|cccc|}
\hline
\rowcolor{mygrey}
& & &
\multicolumn{4}{c|}{Original}
& \multicolumn{4}{c|}{IAAA}
& \multicolumn{4}{c|}{SAAA} \\
\cline{4-15}
\rowcolor{mygrey}
\multirow{-2}{*}{Type} & \multirow{-2}{*}{Method} & \multirow{-2}{*}{Backbone}
& AP & AP$_{50}$ & AP$_{75}$ & AP$_{S}$
& AP & AP$_{50}$ & AP$_{75}$ & AP$_{S}$
& AP & AP$_{50}$ & AP$_{75}$ & AP$_{S}$ \\
\hline\hline
\multirow{6}{*}{Multi-stage}
& Faster R-CNN & R-101
& 49.4 & 76.1 & 54.6 & 31.4
& 51.2 & 77.4 & 58.4 & 28.2
& 50.2 & 76.2 & 55.2 & 33.7 \\
& Cascade R-CNN & R-101
& 50.1 & 76.2 & 54.6 & 32.6
& 52.5 & 78.7 & 60.6 & 31.7
& 52.1 & 75.9 & 56.9 & 34.5 \\
& Grid R-CNN & X-101
& 49.2 & 75.4 & 54.0 & 33.8
& 48.4 & 75.1 & 56.4 & 33.4
& 51.2 & 76.0 & 56.0 & 34.5 \\
& Libra R-CNN & X-101
& 50.3 & 74.5 & 55.2 & 29.7
& 48.4 & 75.9 & 56.4 & 21.4
& 39.9 & 67.3 & 43.4 & 27.8 \\
& RepPoints & R-101
& 44.7 & 75.5 & 48.3 & 30.2
& 47.5 & 76.7 & 84.3 & 24.4
& 48.0 & 77.2 & 52.0 & 33.1 \\
& RepPoints & X-101
& 45.4 & 76.2 & 48.9 & 30.7
& 48.4 & 79.3 & 54.1 & 28.9
& 48.1 & 76.6 & 51.7 & 33.4 \\
\hline
\multirow{5}{*}{One-stage}
& FreeAnchor & R-101
& 42.7 & 71.7 & 44.4 & 21.2
& 47.6 & 73.3 & 54.1 & 19.9
& 48.4 & 73.1 & 52.6 & 29.0 \\
& FSAF & X-101
& 43.4 & 74.0 & 45.8 & 29.8
& 50.3 & 77.9 & 57.6 & 25.5
& 50.4 & 76.8 & 53.5 & 33.9 \\
& ATSS & R-101
& 48.6 & 77.6 & 51.8 & 30.9
& 52.9 & 78.5 & 59.1 & 26.7
& 51.4 & 76.9 & 56.2 & 35.3 \\
& PAA & R-101
& 48.3 & 76.1 & 52.0 & 31.0
& 48.3 & 77.6 & 53.2 & 24.9
& 51.9 & 77.7 & 56.8 & 33.8 \\
& GFL & X-101
& 50.2 & 77.0 & 53.6 & 37.4
& 53.1 & 78.3 & 59.5 & 27.0
& 52.8 & 77.4 & 56.9 & 38.5 \\
\hline
\end{tabular}}
\end{table*}

Because our pipeline integrates an unconditional diffusion model for mask generation and a Pix2Pix-based generator for mask-to-image translation, the system can produce an essentially unlimited number of synthetic samples once trained. To demonstrate this scalability, we expanded the augmentation set to include an additional 500 generated images, as shown in Table~\ref{tab:saaa500}. Under this setting, performance improvements became even more pronounced, reaching +3.35 AP, +0.52 AP$_{50}$, +4.35 AP$_{75}$, and +2.26 AP$_s$ on average. These results illustrate that given sufficient computational resources, our SAAA approach can further boost accuracy by generating larger training sets. So the method can be used in biomedical imaging settings with limited datasets.

\begin{table}[htbp]
\centering
\caption{Detection performance using 500 SAAA-generated images. SAAA continues to improve model performance as the augmented dataset becomes larger.}
\label{tab:saaa500}
\vspace{0.5em}
\renewcommand{\arraystretch}{1.1}
\setlength{\tabcolsep}{4pt}
\fontsize{8pt}{10pt}\selectfont
\begin{tabular}{|l|l|c|cccc|}
\hline
\rowcolor{mygrey}
 & &
& \multicolumn{4}{c|}{SAAA} \\
\cline{4-7}
\rowcolor{mygrey}
\multirow{-2}{*}{Type} & \multirow{-2}{*}{Method} & \multirow{-2}{*}{Backbone} & AP & AP$_{50}$ & AP$_{75}$ & AP$_{S}$ \\
\hline\hline
\multirow{6}{*}{Multi-stage}
& Faster R-CNN & R-101
& 50.9 & 76.1 & 55.3 & 33.8 \\
& Cascade R-CNN & R-101
& 52.6 & 76.1 & 58.3 & 33.3 \\
& Grid R-CNN & X-101
& 52.0 & 75.5 & 56.7 & 35.8 \\
& Libra R-CNN & X-101
& 47.0 & 71.1 & 52.3 & 25.8 \\
& RepPoints & R-101
& 48.7 & 76.1 & 52.7 & 33.8 \\
& RepPoints & X-101
& 49.3 & 77.2 & 53.1 & 34.8 \\
\hline
\multirow{5}{*}{One-stage}
& FreeAnchor & R-101
& 50.1 & 72.9 & 55.2 & 28.2 \\
& FSAF & X-101
& 51.5 & 76.5 & 55.7 & 34.3 \\
& ATSS & R-101
& 51.8 & 77.1 & 56.8 & 34.9 \\
& PAA & R-101
& 53.4 & 77.2 & 59.3 & 35.0 \\
& GFL & X-101
& 53.7 & 78.1 & 57.3 & 38.2 \\
\hline
\end{tabular}
\end{table}

We also evaluated diffusion-based models as image generators, including ControlNet and Stable Diffusion.  However, the results showed that these models performed poorly on medical cell images with simple grayscale appearances and limited color variation in experiment data. These methods generate images through iterative noise prediction. The loss decreased rapidly during training, often reaching low values within the first few iterations. However, this behavior appeared abnormal, likely due to the simplicity of the grayscale input masks. The models may have learned to denoise trivial patterns rather than capturing meaningful structures.
 What's more, the resulting images aligned poorly with biological morphology, often exhibiting unrealistic textures, saturated colors, and over-stylized visual artifacts. Especially when using pre-trained weights, the generated outputs deviated further from the expected cell structures. This is likely caused by domain mismatch and the complex priors embedded in large-scale natural image training. Such characteristics make diffusion models less suitable for biomedical tasks that require semantic precision and anatomical consistency. Overall, diffusion-based models appear more suitable for complex natural scenes rather than biomedical applications, which require semantic precision and anatomical consistency.

In contrast, our Pix2Pix-based generator achieved better semantic control and visual quality, especially in settings with limited cell data and object structures.
In addition, the overall pipeline is computationally efficient. Training the Pix2Pix model on a single GPU takes less than 6 hours, and image generation takes less than 0.1 seconds per sample. Compared to diffusion models, this provides significant speed-up and makes the method practical for large-scale or resource-limited environments.

\section{Discussion}\label{sec:discussion}

We presented Instance Aware Automatic Augmentation (IAAA) for the generation of artificial cell nuclei microscopical images along with their correct instance segmentation masks. An initial set of segmentation objects is used with Greedy AutoAugment to find the best-performing policies. The found policies and initial set of segmentation objects are then used to create the final artificial images. The images are compared with state-of-the-art data augmentation methods. The results show that the quality of the proposed method in different stages is on par with the original data and is competitive compared to different GAN models. In parallel, to increase our data generation capacity, we introduced the Semantic-Aware AI Augmentation (SAAA) pipeline. It uses a diffusion model to create masks and a Pix2Pix network to synthesize images. This approach produces a very large number of realistic samples and expands the training set far beyond what non-AI augmentation methods can provide, while preserving the distributional characteristics and semantic information of real cells. These observations are confirmed by FID and KID scores. Overall, when the number of augmented samples kept the same, IAAA and SAAA produce similar improvements. Nonetheless, SAAA provides superior scalability. With sufficient computational resources, it can generate larger datasets and offers a more advantageous option for further performance gains. The experiments demonstrate that the proposed augmentation technique consistently improves the detection of cells which is confirmed by variants of AP scores for both bounding box detection and instance segmentation. The proposed method is effectively ready to generate artificial cell nuclei microscopical images. In the future, this method can help to better train deep learning models on a broader range of microscopical cell images and related biomedical datasets.

%% \section*{Declaration of competing interest}
%%
%% The authors declare that they have no known competing financial interests or personal relationships that could have appeared to influence the work reported in this paper.

%% \section*{Data availability}
%%
%% Data will be made available on request.

\section*{Acknowledgements}

%% Not applicable.
%%
%% \section*{Funding}

This work was supported in part by AI130197 (DL-Rutgers), New Jersey Health Foundation, ScaleReady’s G-Rex Grant, and HealthAdvance Award (part of U01HL150852). Research reported in this publication was supported by the National Institute Of Allergy And Infectious Diseases of the National Institutes of Health under Award Number R01AI130197. The content is solely the responsibility of the authors and does not necessarily represent the official views of the National Institutes of Health.

%% CARNA  CAR-T carn-k
%% 
%% SUPPPLEMENT

%% \section*{CRediT authorship contribution statement}
%%
%% \textbf{Xiang Zhang}: Conceptualization, Methodology, Software, Validation, Formal analysis, Investigation, Data curation, Writing, Visualization.
%% \textbf{Boxuan Zhang}: Methodology, Software, Validation.
%% \textbf{Alireza Naghizadeh}: Methodology, Resources.
%% \textbf{Mohab Mohamed}: Resources, Data curation.
%% \textbf{Dongfang Liu (RIT)}: Supervision, Project administration.
%% \textbf{Ruixiang Tang}: Supervision, Writing (review and editing).
%% \textbf{Dimitris Metaxas}: Supervision, Writing (review and editing).
%% \textbf{Dongfang Liu (Rutgers)}: Conceptualization, Supervision, Project administration, Writing (review and editing).

%% References
\bibliographystyle{elsarticle-num}
\bibliography{reference}

\appendix

\section{Supplementary Materials}

\subsection{Introduction in the Supplementary Data}
In IAAA, an initial set of images is manually segmented and then sorted between two categories, the first being single-cell objects (SCOs), which contain singular cells that can be easily distinguished from other cells, and the second type being multi-cell objects (MCOs) which contain multiple cells and more commonly suffer from impurities such as low contrast of cell boundaries, background noise, adhesion, and cell clustering.

When these two categories are used to generate images, we create artificial images that, for the most part, are not distinguishable from natural data. Specifically, normal augmentation techniques are applied to the segmented SCOs and MCOs. To maintain the integrity of the original cells, only a subset of controlled augmentation techniques is used. To apply an augmentation onto the cells, we use an augmentation policy. An augmentation policy consists of a set of different techniques at varying magnitudes. Search algorithms, such as Greedy AutoAugment~\cite{naghizadeh2020greedy,peng2018jointly}, find the strongest augmentation policies in the search space of policies using a specified scoring criterion.

A high-scoring augmentation policy transforms an image while maintaining its authentic properties, avoiding obscure augmentations. To find the best policies, a Wasserstein autoencoder~\cite{tolstikhin2017wasserstein,gulrajani2017improved,lindernoren2018pytorchgan} provides a scoring system for Greedy AutoAugment~\cite{naghizadeh2020greedy} which ranks different combinations of sub-policies. The autoencoder attempts to learn an identity function that maps an object to itself, minimizing reconstruction error between inputs and its respective outputs~\cite{bank2023autoencoders}. The features found through this neural network help to retain a level of authenticity when manipulating an object. The best-performing sub-policies are then used as final policies in creating training images.

An image is generated by randomly selecting the number of cells within an image and applying a policy for each cell. The same procedure is used for the available masked images to create artificial images with complete ground truth masks. The cells need to be placed on the background frames, which have similar textures to the original images. To support images for datasets with backgrounds of bright color spectrums, we propose solutions to generate artificial backgrounds and the placements of cells on these generated backgrounds. In both processes, we want the images to look as natural as possible and follow the patterns of the real datasets.

To assess the quality of the generated images, we follow a similar approach found in Chong and Forsyth~\cite{chong2020effectively}. To evaluate the quality of the images, we use the Frechet Inception Distance (FID) as well as the Kernel Inception Distance (KID), which has shown better support for smaller datasets~\cite{binkowski2018demystifying}.

To measure the accuracy of cell detection, we use variants of average precision (AP). The experiments demonstrate that the proposed augmentation technique generates realistic artificial images and consistently improves the detection and segmentation of nuclei microscopical images.

\subsection{Supplementary Tables}
\setcounter{table}{0}
\renewcommand{\thetable}{S\arabic{table}}

\begin{table*}[htbp]
  \centering
  \caption{Supplementary Table S1: The list of selected augmentation techniques used in the IAAA method.}
  \label{tab:s1}
  \vspace{0.5em}
  \resizebox{\textwidth}{!}{
  \begin{tabular}{llp{0.28\textwidth}llp{0.28\textwidth}}
    \hline
    \rowcolor{mygrey}
    No. & Technique & Description & No. & Technique & Description \\
    \hline
    1 & FlipLR & Flipping the image along the vertical axis. & 7 & Contrast & Changing the contrast of the image. \\
    2 & FlipUD & Flipping the image along the horizontal axis. & 8 & Brightness & Adjusting the brightness of the image. \\
    3 & AutoContrast & Increasing the contrast of the image. & 9 & Sharpness & Adjusting the sharpness of the image. \\
    4 & Equalize & Equalizing the histogram of the image. & 10 & Smooth & Smoothing the image (low-pass filtering). \\
    5 & Rotate & Rotating the image by certain degrees. & 11 & Resize & Changes the image resolution. \\
    6 & Posterize & Reducing the number of bits for each pixel. &  &  &  \\
    \hline
  \end{tabular}}
\end{table*}

\begin{table*}[htbp]
  \centering
  \caption{Supplementary Table S2: The FID and KID scores of SCOs, MCOs, and final cell image generation using the Kaggle dataset.}
  \label{tab:s2}
  \resizebox{\textwidth}{!}{
  \begin{tabular}{lcccccc}
    \hline
    \rowcolor{mygrey}
    Method & SCO FID & SCO KID & MCO FID & MCO KID & Final FID & Final KID \\
    \hline
    DCGAN & 235.60 & 0.2042 & 404.90 & 0.2986 & 435.55 & 0.3352 \\
    BigGAN-SD & 363.57 & 0.3320 & 414.81 & 0.3409 & 496.38 & 0.5216 \\
    BigGAN-Diff & 262.46 & 0.4875 & 126.30 & 0.1313 & 297.84 & 0.3037 \\
    StyleGAN2-Diff & 10.36 & 0.0067 & 99.04 & 0.0999 & 196.92 & 0.0892 \\
    IAAA (ours) & 23.32 & 0.0212 & 39.93 & 0.0732 & 102.99 & 0.0716 \\
    \hline
  \end{tabular}}
\end{table*}

\subsection{Kaggle Dataset Experiments}
\noindent\textbf{Data Preparation for Original Datasets.} The Kaggle 2018 Data Science Bowl contains 589 images that were acquired under a variety of conditions and vary in cell type, magnification, and imaging modality (brightfield vs. fluorescence)~\cite{caicedo2019nucleus}. A subset of 193 images is selected, which share consistent attributes, and are divided into train, test, and validation sets with 117, 38, and 38 images.

\noindent\textbf{Data Preparation for Mixed Datasets.} The training set for the Kaggle 2018 Data Science Bowl dataset contains 200 images which is a combination of 117 original images and 83 generated images.

\noindent\textbf{SCO Image Generation.} For Frechet Inception Distance, when using the Kaggle dataset, StyleGAN2-Diff scores 10.36, which is marginally better than the proposed method. For Kernel Inception Distance, when using the Kaggle dataset, StyleGAN2-Diff scores 0.0067, which is slightly better than the proposed method.

\noindent\textbf{MCO Image Generation.} The Frechet Inception Distance score of the proposed method surpasses the best-performing GAN model StyleGAN2-Diff by 59.11 within the Kaggle dataset. For the KID, the results remain consistent with the FID values, with the proposed method outperforming StyleGAN2 by 0.0267 using the Kaggle dataset.

\noindent\textbf{Final Image Generation.} The proposed method consistently outperforms the best-performing GAN method, StyleGAN2, by 93.93. The results remain consistent with the KID values, with the proposed method outperforming StyleGAN2 by 0.0176 using the Kaggle dataset.

\noindent\textbf{Performance of Bounding Box Detection.} We present the effectiveness of the proposed augmentation method on the Kaggle dataset in the BBox evaluation sections of Supplementary Table~\ref{tab:s3}. According to the results, we additionally observe that the augmented data consistently improves the results when using the Kaggle dataset. For AP, 10 out of the 11 methods show improvements. AP$_{50}$ improves for 6 out of the 11 methods, AP$_{75}$ improves for 10 out of the 11 methods, and AP$_s$ shows a performance increase across 10 out of the 11 methods. Overall, the improvements to the 11 methods are consistent. The average performance gain is 3.38\% across the experiments.

\begin{table*}[htbp]
  \centering
    \fontsize{8pt}{10pt}\selectfont
  \caption{Supplementary Table S3: The AP, AP$_{50}$, AP$_{75}$, and AP$_s$ scores of box detection on the Kaggle dataset using the IAAA method. The values for Datasets represent the original images, and the values for AUG + Datasets represent the original images with the addition of augmented images. For instance segmentation accuracy, we use KG Instance Segmentation~\cite{yi2019multiscale}.}
  \label{tab:s3}
  \resizebox{\textwidth}{!}{
  \begin{tabular}{llcccccccc}
    \hline
    \rowcolor{mygrey}
     &  & \multicolumn{4}{c}{Datasets} & \multicolumn{4}{c}{AUG + Datasets} \\
    \rowcolor{mygrey}
    \multirow{-2}{*}{Method} & \multirow{-2}{*}{Backbone} & AP & AP$_{50}$ & AP$_{75}$ & AP$_s$ & AP & AP$_{50}$ & AP$_{75}$ & AP$_s$ \\
    \hline
    \multicolumn{10}{l}{Multi-stage BBox Evaluation:} \\
    Faster R-CNN & R-101 & 38.6 & 86.3 & 23.8 & 38.6 & 44.0 & 87.4 & 39.4 & 44.0 \\
    Cascade R-CNN & R-101 & 37.5 & 86.5 & 20.5 & 37.5 & 37.2 & 82.4 & 27.2 & 37.2 \\
    Grid R-CNN & X-101 & 35.0 & 83.3 & 18.2 & 35.0 & 40.2 & 82.8 & 33.4 & 40.2 \\
    Libra R-CNN & X-101 & 43.8 & 88.2 & 34.7 & 43.8 & 44.5 & 88.8 & 36.8 & 44.5 \\
    RepPoints & R-101 & 38.8 & 86.9 & 24.4 & 38.8 & 40.0 & 82.1 & 26.1 & 40.0 \\
    RepPoints & X-101 & 35.8 & 86.7 & 12.7 & 35.9 & 38.9 & 86.8 & 24.7 & 38.9 \\
    \multicolumn{10}{l}{One-stage BBox Evaluation:} \\
    FreeAnchor & R-101 & 33.3 & 82.8 & 16.7 & 33.3 & 33.4 & 81.4 & 16.1 & 33.4 \\
    FSAF & X-101 & 37.8 & 86.1 & 22.5 & 37.8 & 40.6 & 87.0 & 30.7 & 40.6 \\
    ATSS & R-101 & 36.3 & 83.3 & 19.0 & 36.4 & 37.2 & 84.1 & 21.2 & 37.2 \\
    PAA & R-101 & 38.5 & 85.5 & 23.7 & 38.5 & 46.4 & 82.0 & 47.1 & 46.6 \\
    GFL & X-101 & 37.1 & 88.6 & 15.2 & 37.1 & 40.4 & 89.2 & 27.3 & 40.4 \\
    \hline
  \end{tabular}}

  \vspace{0.5em}

  \begin{tabular}{lcccc}
    \hline
    \rowcolor{mygrey}
    Metric System & AP@0.5 (Kaggle) & IoU@0.5 (Kaggle) & AP@0.7 (Kaggle) & IoU@0.7 (Kaggle) \\
    \hline
    Mixed Dataset & 70.71 & 80.30 & 51.25 & 84.79 \\
    Original Dataset & 62.44 & 79.72 & 50.76 & 84.56 \\
    \hline
  \end{tabular}
\end{table*}

\subsection{Generated Image Samples}
In this section, we present the additional image samples at higher resolutions. The higher resolutions help to compare the images from real samples and the ones that are generated by the IAAA method. Supplementary Figures S1--S6 include images with zoomed-in areas to show SCOs and MCOs for Neural, CAR-T/NK, and Kaggle datasets. Supplementary Figures S7--S10 show SCOs and MCOs without their placements on the frames for both real samples and generated images. Supplementary Figures S11--S12 show the effect of applying different forms of augmentation policies on SCOs and MCOs. Supplementary Figures S13--S19 show the final images of StyleGAN2-Diff, IAAA, and real samples. The images from StyleGAN2-Diff are presented because StyleGAN2-Diff had the best quality among GAN models.

\section{Supplementary Figures}
\setcounter{figure}{0}
\renewcommand{\thefigure}{S\arabic{figure}}

\begin{figure}[htbp]
  \centering
  \includegraphics[width=\linewidth]{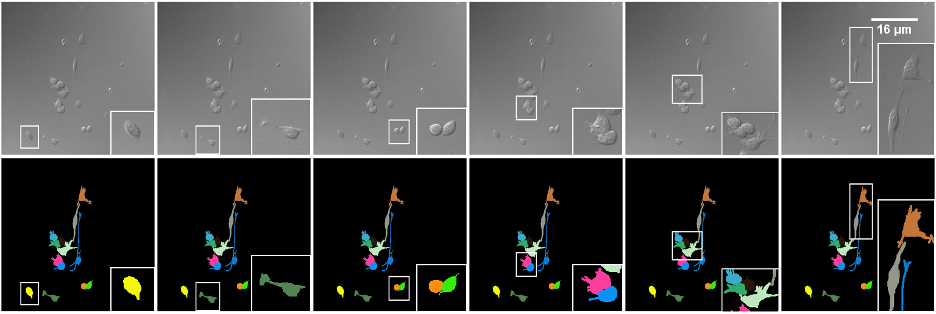}
  \caption{A sample of an artificially generated Neural dataset (first row) with its respective instance segmentation (second row). In six different locations, the zoom-in functionality is used to provide better details of the generated cells and their respective masks.}
  \label{fig:s1}
\end{figure}

\begin{figure}[htbp]
  \centering
  \includegraphics[width=\linewidth]{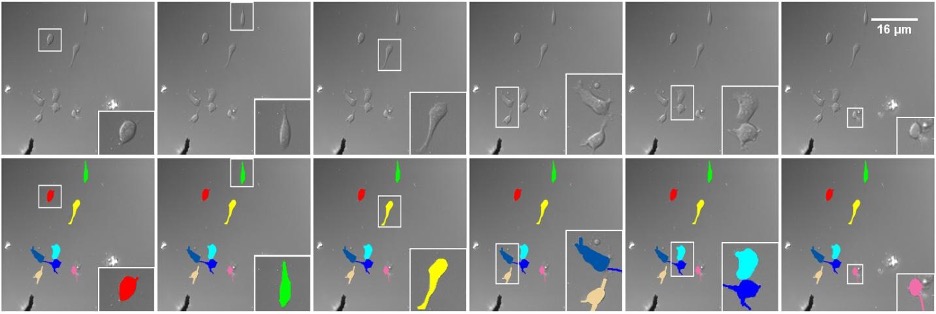}
  \caption{A real sample from the Neural dataset (first row) with its respective instance segmentation (second row). In six different locations, the zoom-in functionality is used to provide better details of the real cells and their respective masks.}
  \label{fig:s2}
\end{figure}

\begin{figure}[htbp]
  \centering
  \includegraphics[width=\linewidth]{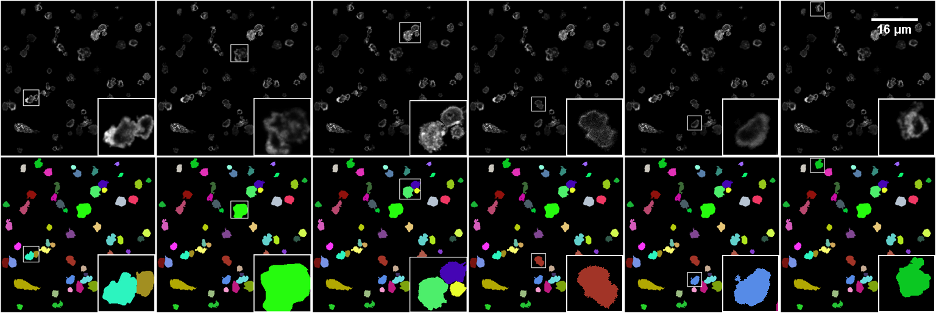}
  \caption{A sample of an artificially generated CAR-T/NK dataset (first row) with its respective instance segmentation (second row). In six different locations, the zoom-in functionality is used to provide better details of the generated cells and their respective masks.}
  \label{fig:s3}
\end{figure}

\begin{figure}[htbp]
  \centering
  \includegraphics[width=\linewidth]{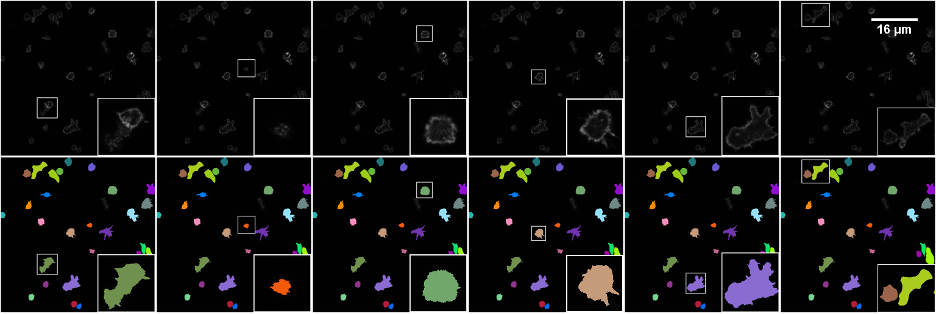}
  \caption{A real sample from the CAR-T/NK dataset (first row) with its respective instance segmentation (second row). In six different locations, the zoom-in functionality is used to provide better details of the real cells and their respective masks.}
  \label{fig:s4}
\end{figure}

\begin{figure}[htbp]
  \centering
  \includegraphics[width=\linewidth]{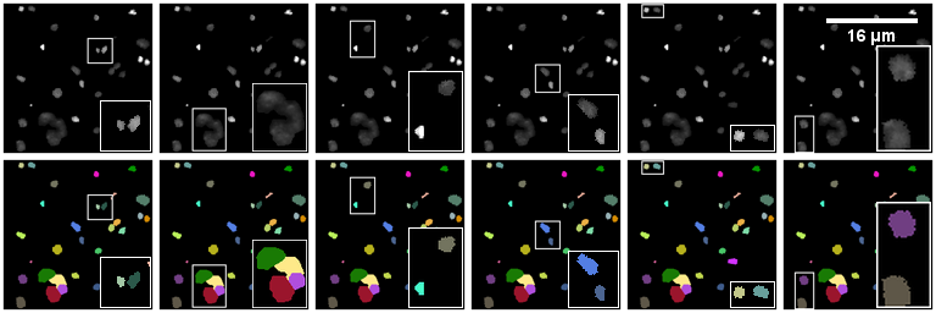}
  \caption{A sample of an artificially generated Kaggle dataset (first row) with its respective instance segmentation (second row). In six different locations, the zoom-in functionality is used to provide better details of the generated cells and their respective masks.}
  \label{fig:s5}
\end{figure}

\begin{figure}[htbp]
  \centering
  \includegraphics[width=\linewidth]{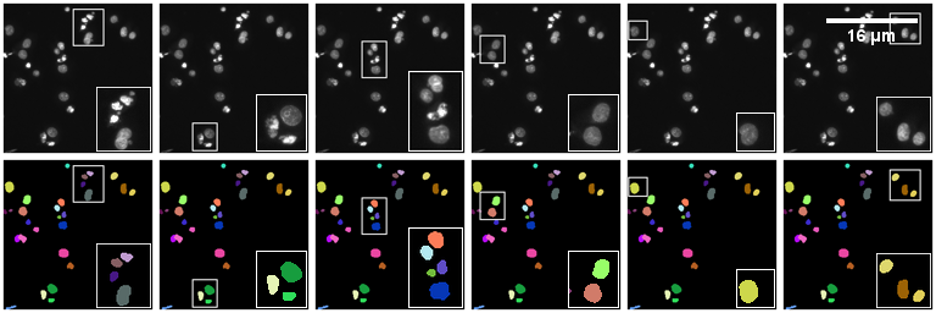}
  \caption{A real sample from the Kaggle dataset (first row) with its respective instance segmentation (second row). In six different locations, the zoom-in functionality is used to provide better details of the real cells and their respective masks.}
  \label{fig:s6}
\end{figure}

\begin{figure}[htbp]
  \centering
  \includegraphics[width=\linewidth]{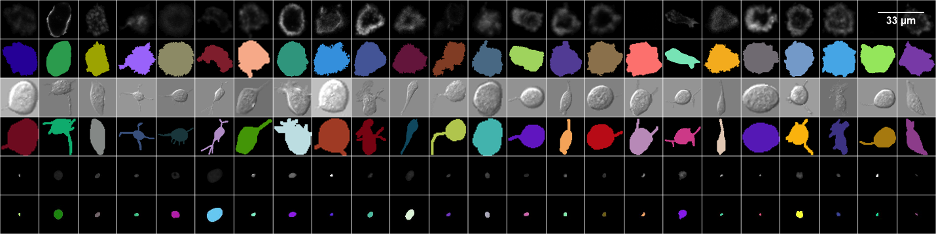}
  \caption{SCO real images on CAR-T/NK, Neural, and Kaggle datasets; the first two rows represent the SCO images and their masks from the CAR-T/NK dataset. The third and fourth rows represent the SCO images and their masks from the Neural dataset. The fifth and sixth rows represent the SCO images and their masks from the Kaggle dataset.}
  \label{fig:s7}
\end{figure}

\begin{figure}[htbp]
  \centering
  \includegraphics[width=\linewidth]{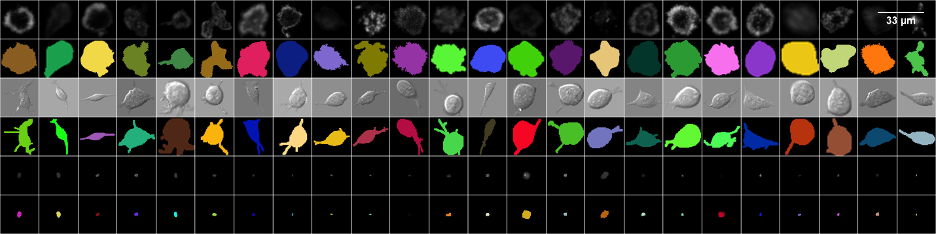}
  \caption{SCO image generation on CAR-T/NK, Neural, and Kaggle datasets; the first two rows represent the SCO images and their masks from the CAR-T/NK dataset. The third and fourth rows represent the SCO images and their masks from the Neural dataset. The fifth and sixth rows represent the SCO images and their masks from the Kaggle dataset.}
  \label{fig:s8}
\end{figure}

\begin{figure}[htbp]
  \centering
  \includegraphics[width=\linewidth]{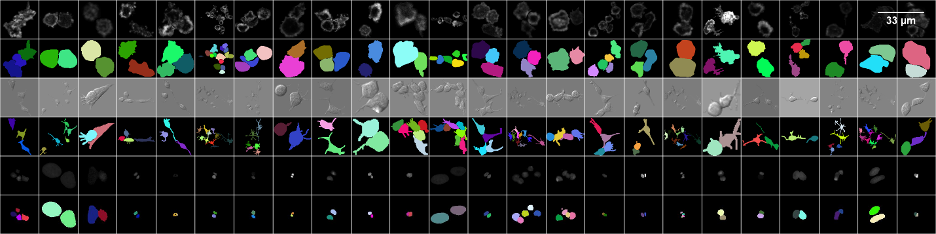}
  \caption{MCO real images on CAR-T/NK, Neural, and Kaggle datasets; the first two rows represent the MCO images and their masks from the CAR-T/NK dataset. The third and fourth rows represent the MCO images and their masks from the Neural dataset. The fifth and sixth rows represent the MCO images and their masks from the Kaggle dataset.}
  \label{fig:s9}
\end{figure}

\begin{figure}[htbp]
  \centering
  \includegraphics[width=\linewidth]{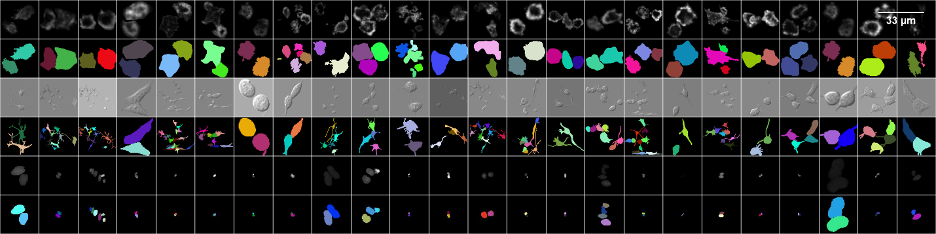}
  \caption{MCO image generation on CAR-T/NK, Neural, and Kaggle datasets; the first two rows represent the MCO images and their masks from the CAR-T/NK dataset. The third and fourth rows represent the MCO images and their masks from the Neural dataset. The fifth and sixth rows represent the MCO images and their masks from the Kaggle dataset.}
  \label{fig:s10}
\end{figure}

\begin{figure}[htbp]
  \centering
  \includegraphics[width=\linewidth]{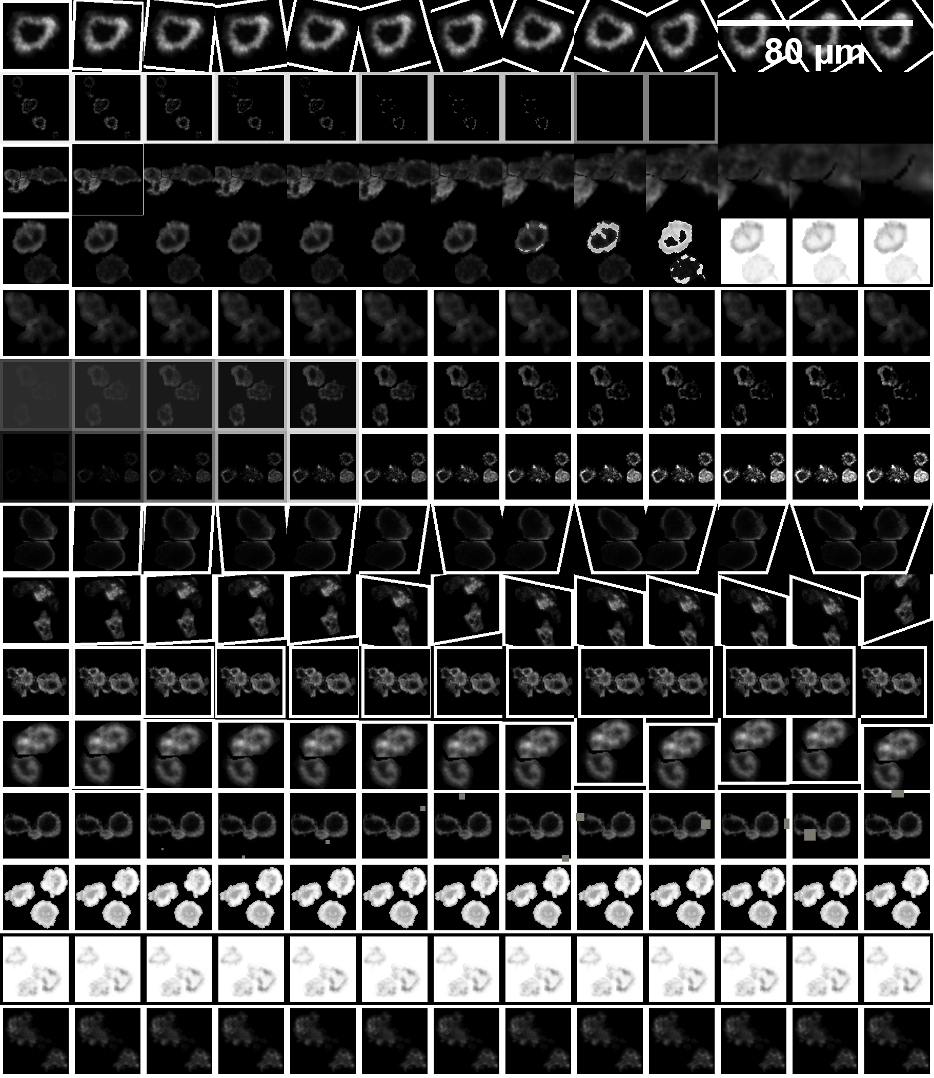}
  \caption{A sample using different forms of augmentation techniques on the cells. Each row represents a new augmentation technique applied on a MCO within the CAR-T/NK dataset. The augmentations used in the order of rows are: 'Rotate', 'Posterize', 'CropBilinear', 'Solarize', 'Color', 'Contrast', 'Brightness', 'ShearX', 'ShearY', 'TranslateX', 'TranslateY', 'Cutout', 'Equalize', 'Invert', 'AutoContrast'.}
  \label{fig:s11}
\end{figure}

\begin{figure}[htbp]
  \centering
  \includegraphics[width=\linewidth]{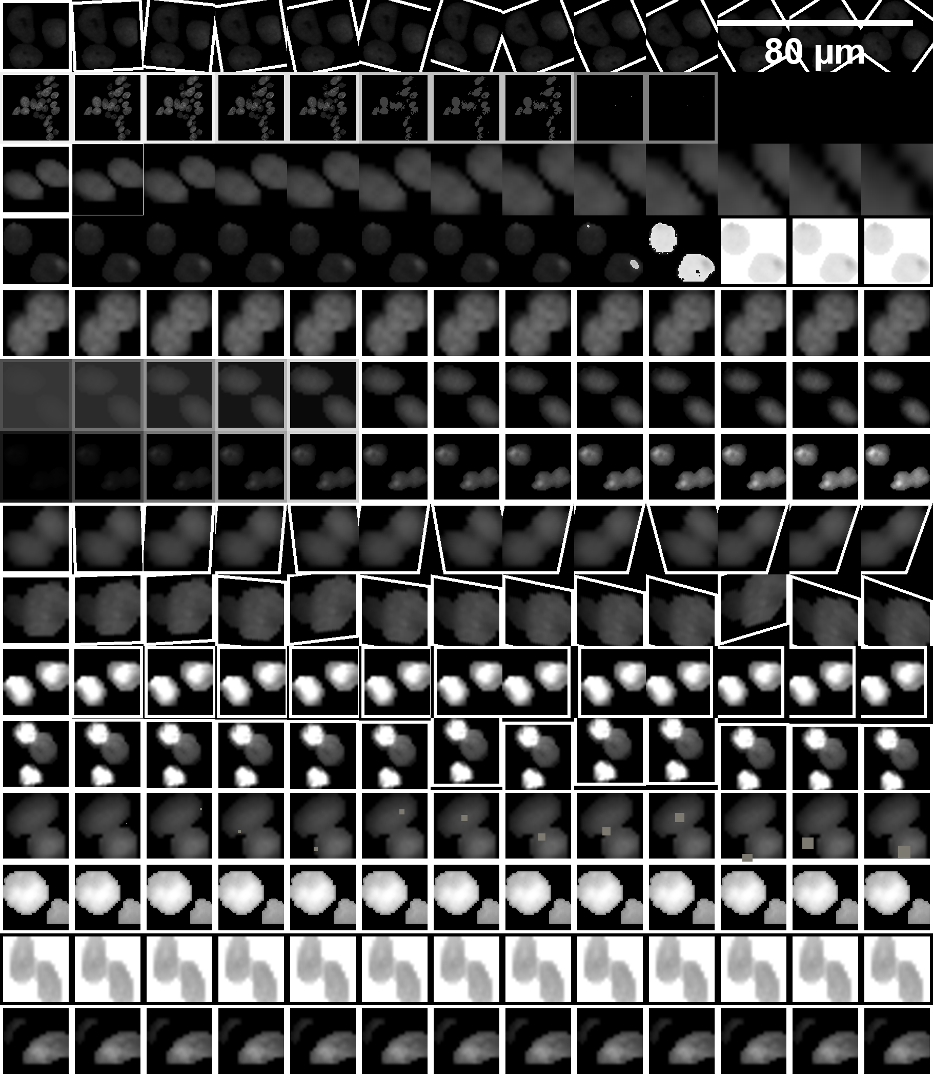}
  \caption{A sample using different forms of augmentation techniques on the cells. Each row represents a new augmentation technique applied on a MCO within the Kaggle dataset. The augmentations used in the order of rows are: 'Rotate', 'Posterize', 'CropBilinear', 'Solarize', 'Color', 'Contrast', 'Brightness', 'ShearX', 'ShearY', 'TranslateX', 'TranslateY', 'Cutout', 'Equalize', 'Invert', 'AutoContrast'.}
  \label{fig:s12}
\end{figure}

\begin{figure}[htbp]
  \centering
  \includegraphics[width=\linewidth]{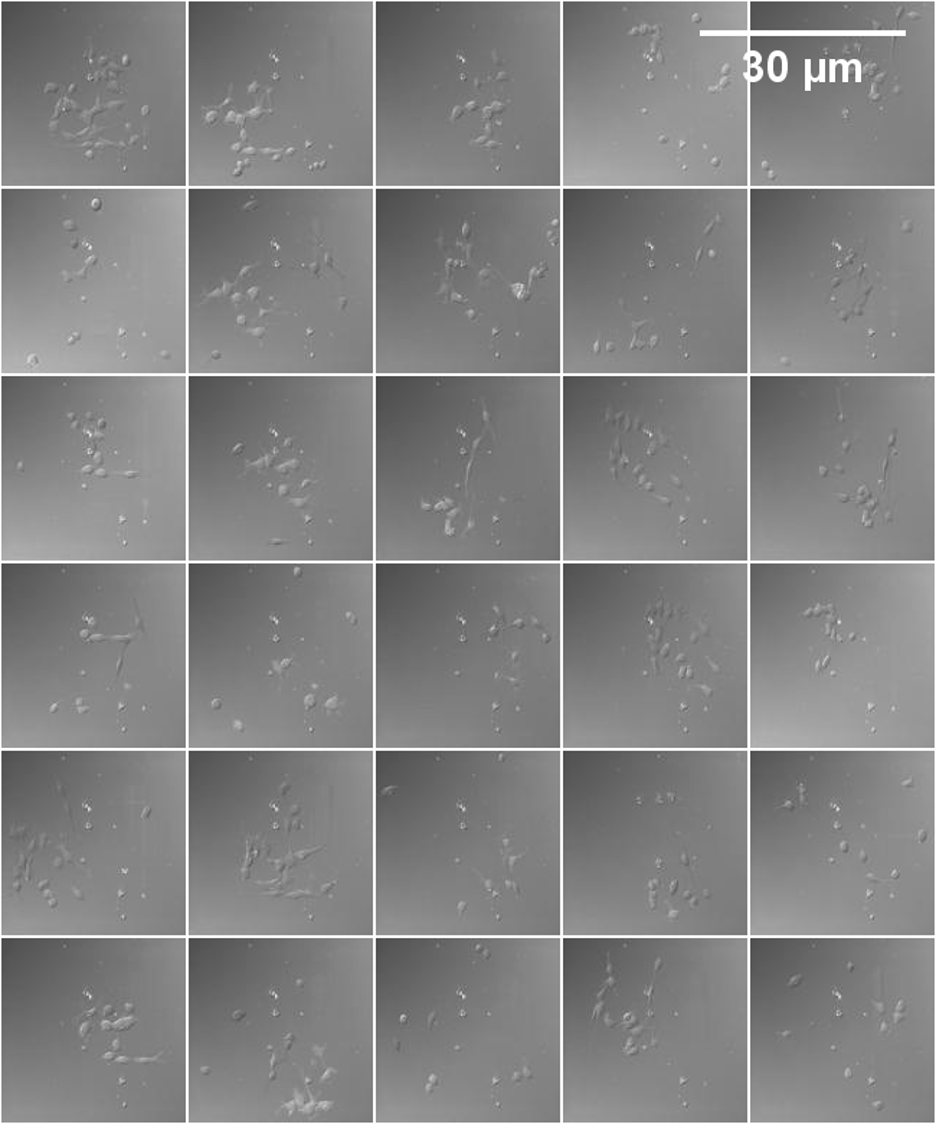}
  \caption{A sample of the final images created from the proposed method (128$\times$128 for each sample) using the Neural dataset.}
  \label{fig:s13}
\end{figure}

\begin{figure}[htbp]
  \centering
  \includegraphics[width=\linewidth]{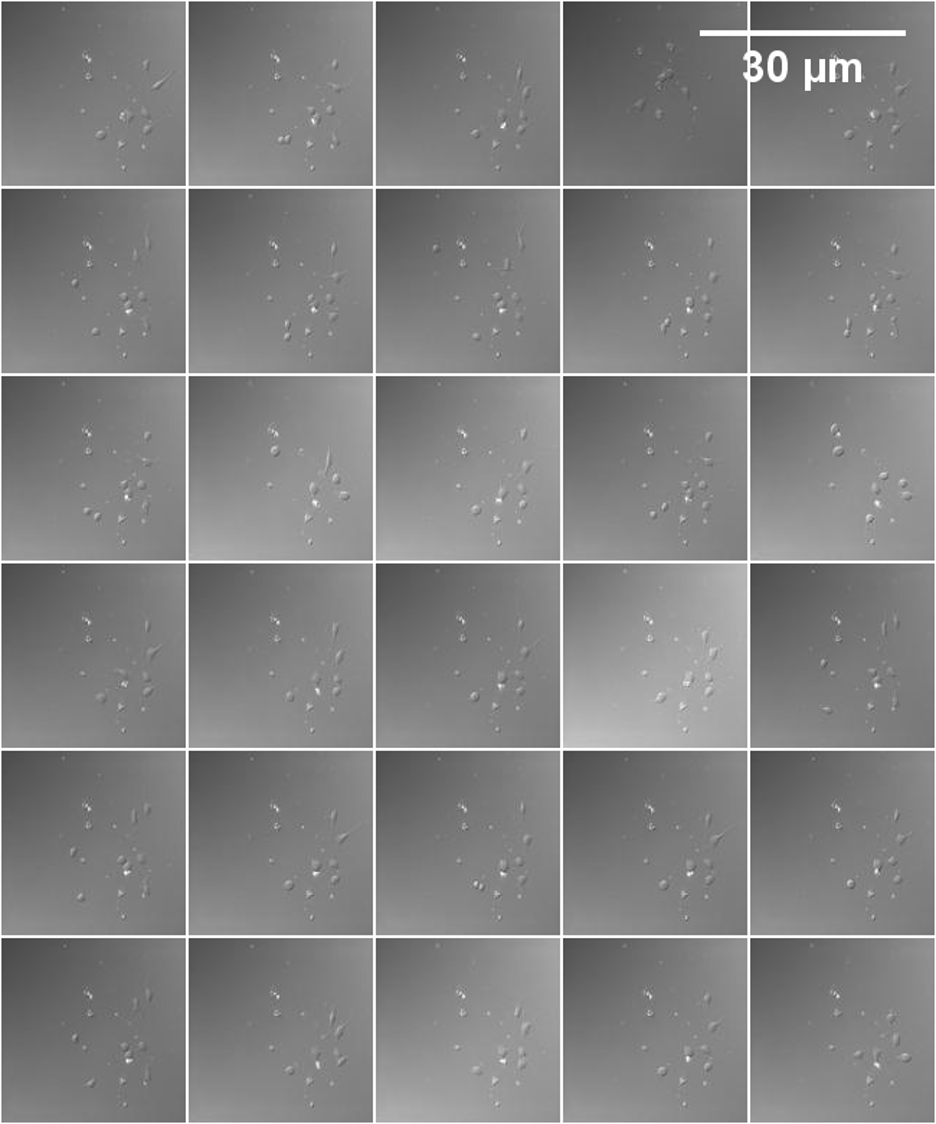}
  \caption{A sample of the final images created from the real images (128$\times$128 for each sample) using the Neural dataset.}
  \label{fig:s14}
\end{figure}

\begin{figure}[htbp]
  \centering
  \includegraphics[width=\linewidth]{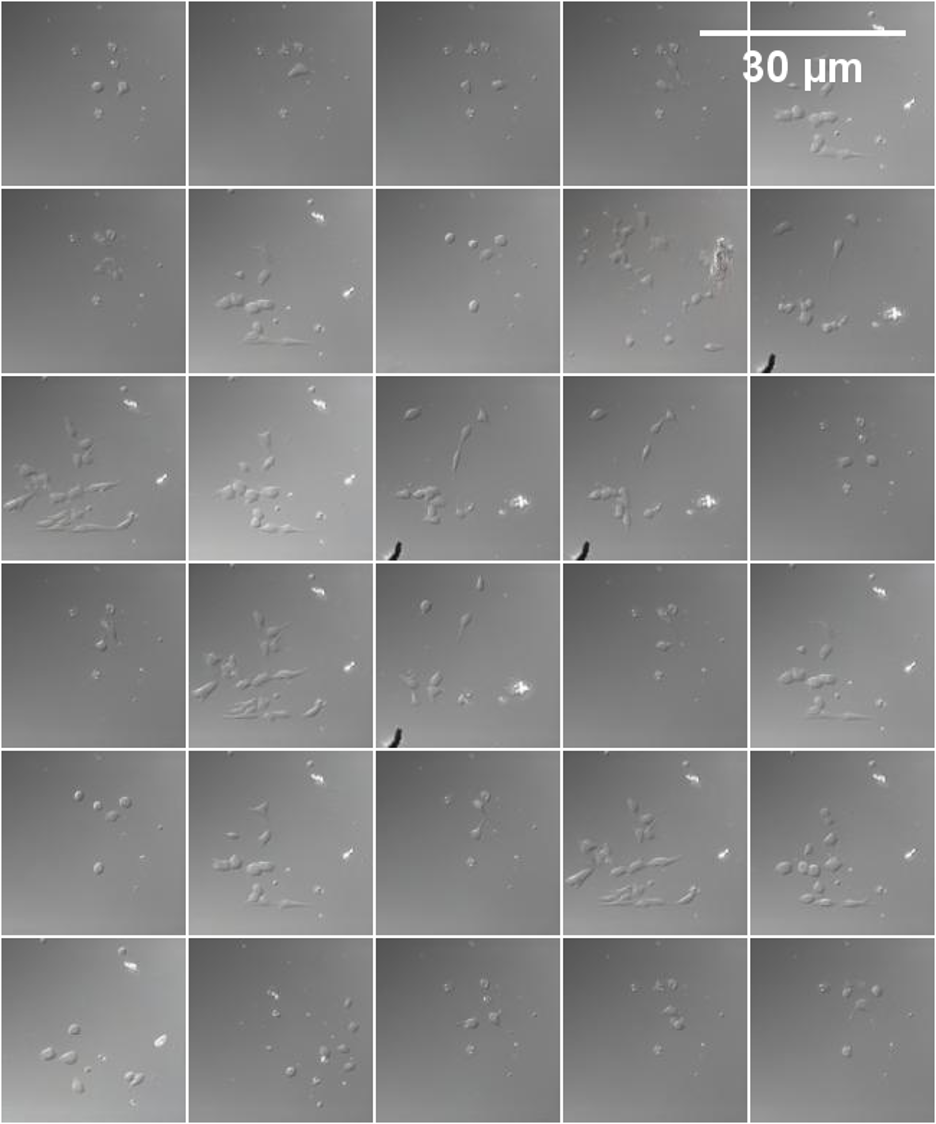}
  \caption{A sample of the final images created from the StyleGAN2-Diff method (128$\times$128 for each sample) using the Neural dataset.}
  \label{fig:s15}
\end{figure}

\begin{figure}[htbp]
  \centering
  \includegraphics[width=\linewidth]{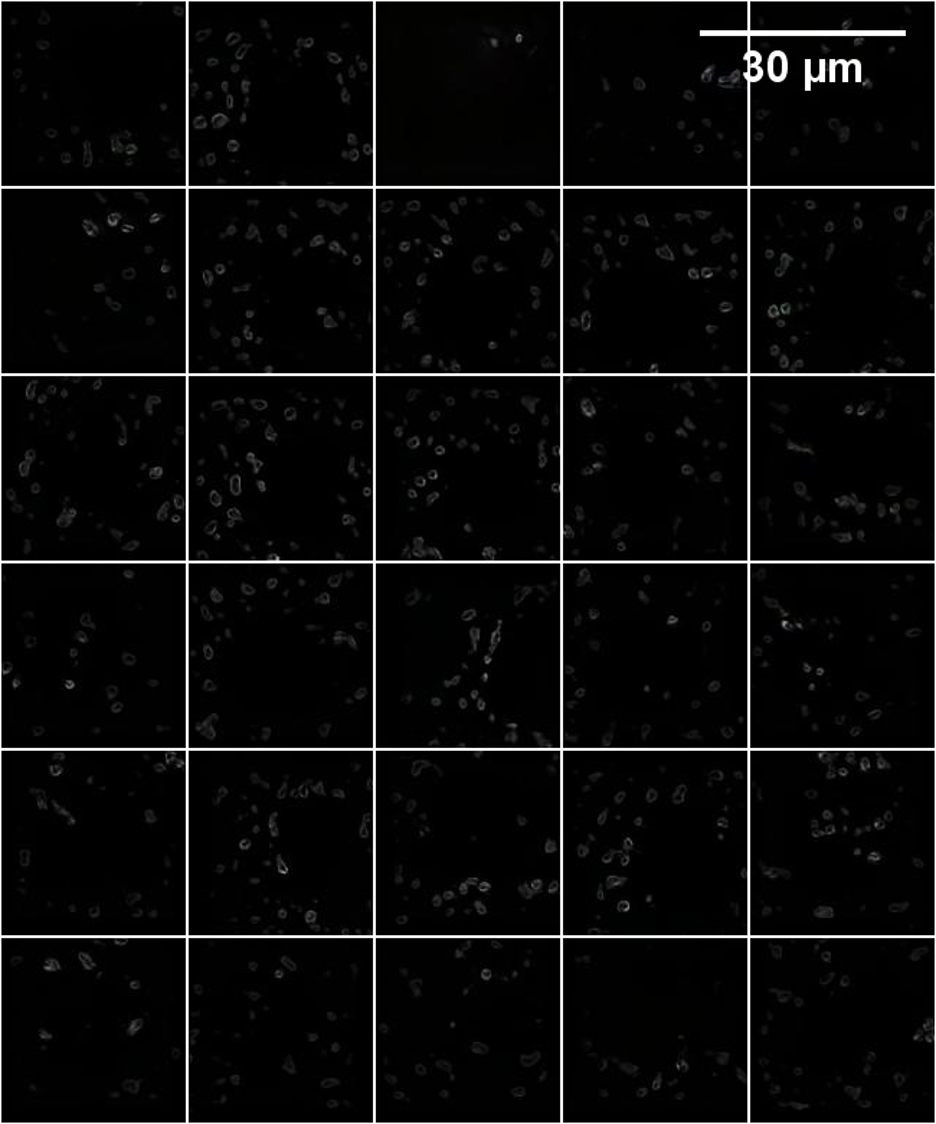}
  \caption{A sample of the final images created from the StyleGAN2-Diff method (128$\times$128 for each sample) using the CAR-T/NK dataset.}
  \label{fig:s16}
\end{figure}

\begin{figure}[htbp]
  \centering
  \includegraphics[width=\linewidth]{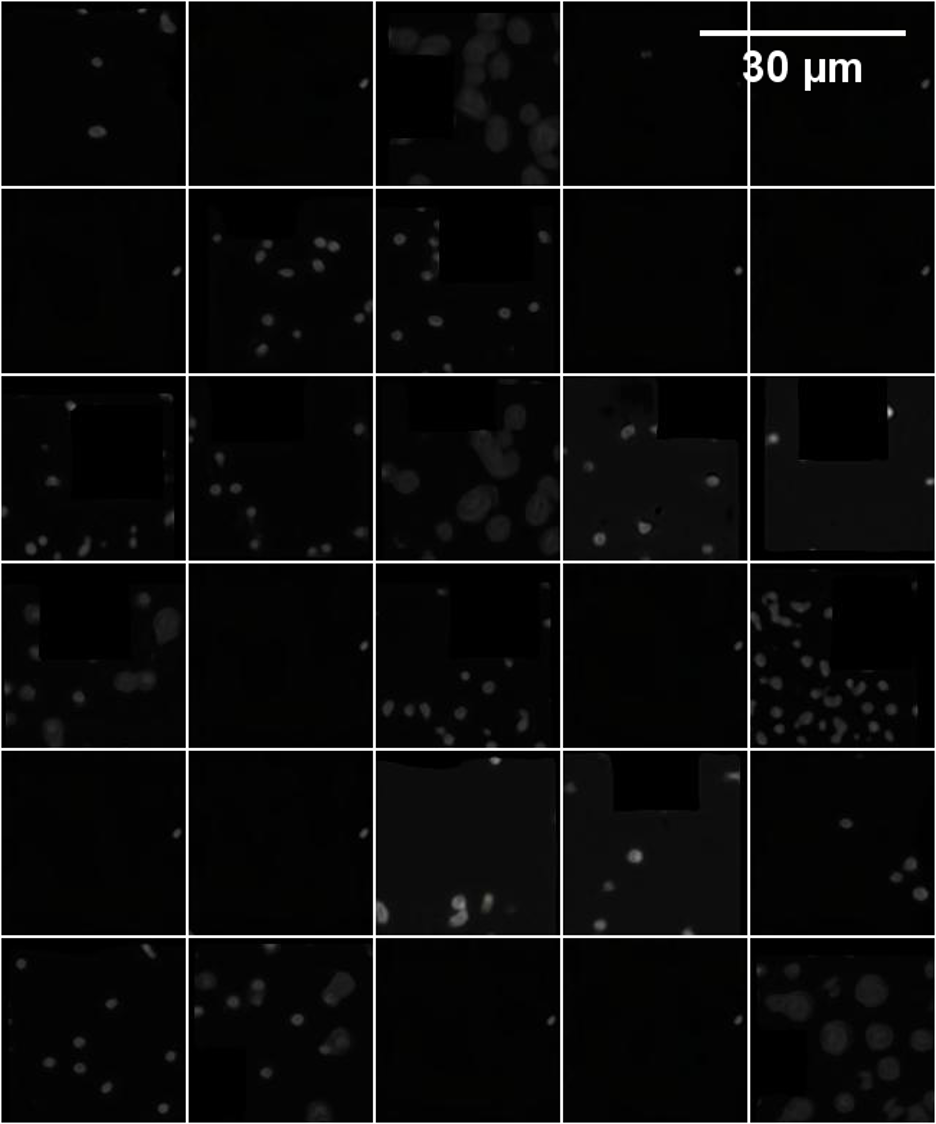}
  \caption{A sample of the final images created from the StyleGAN2-Diff method (128$\times$128 for each sample) using the Kaggle dataset.}
  \label{fig:s17}
\end{figure}

\begin{figure}[htbp]
  \centering
  \includegraphics[width=\linewidth]{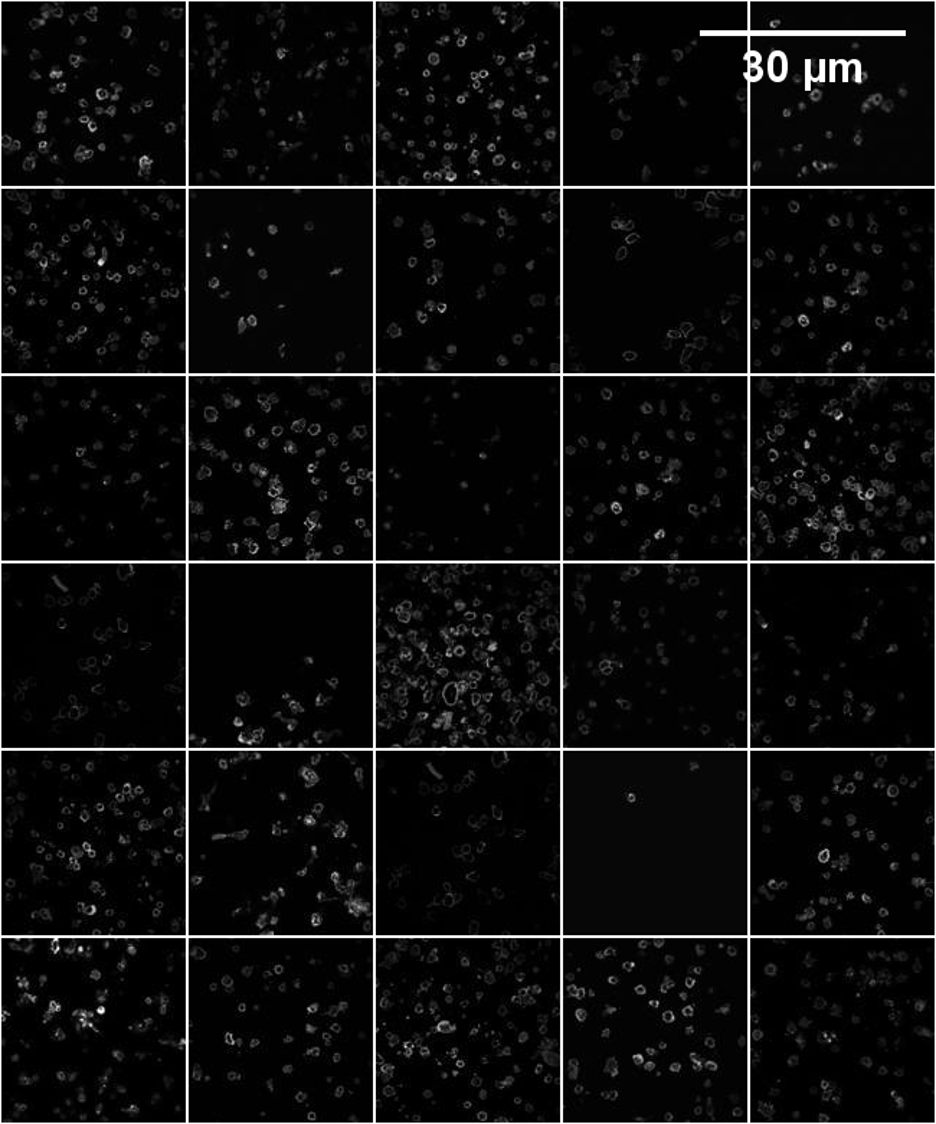}
  \caption{A sample of the final images created from the real images (128$\times$128 for each sample) using the CAR-T/NK dataset.}
  \label{fig:s18}
\end{figure}

\begin{figure}[htbp]
  \centering
  \includegraphics[width=\linewidth]{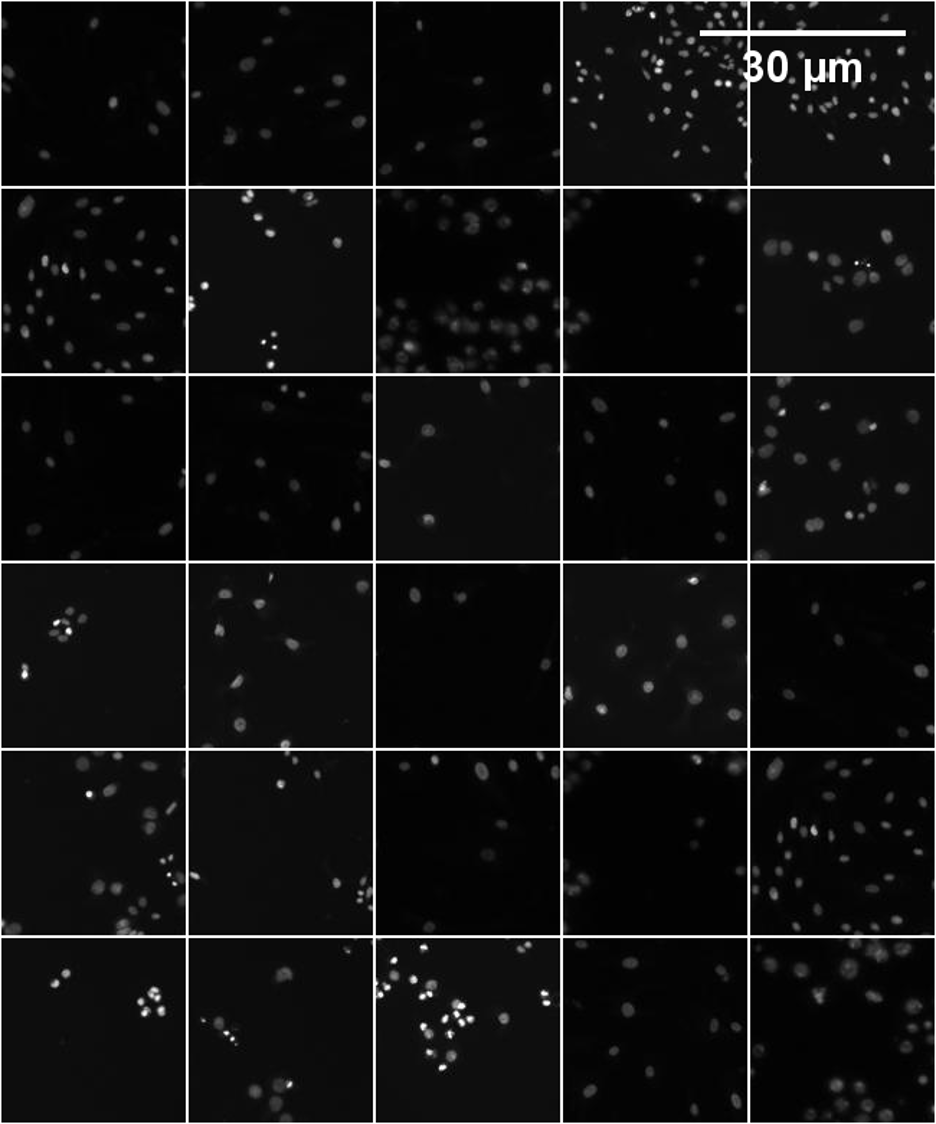}
  \caption{A sample of the final images created from the real images (128$\times$128 for each sample) using the Kaggle dataset.}
  \label{fig:s19}
\end{figure}

\newpage
\section*{Declaration of competing interest}

The authors declare that they have no known competing financial interests or personal relationships that could have appeared to influence the work reported in this paper.

\end{document}